%% file: arxiv_main.tex
\documentclass{article}

\usepackage{microtype}
\usepackage{graphicx}
\usepackage{subfigure}
\usepackage{booktabs}
\usepackage{colortbl}
\usepackage{mathtools}
\usepackage{nicefrac}
\usepackage{enumitem}
\usepackage{hyperref}

\input{math_commands.tex}

\renewcommand{\method}{\textup{\textsc{VerKV}}}
\let\origincludegraphics\includegraphics
\renewcommand{\includegraphics}[2][]{\IfFileExists{arxiv_figs/#2}{\origincludegraphics[#1]{arxiv_figs/#2}}{\origincludegraphics[#1]{#2}}}

\usepackage[accepted]{mlsys2025}
\usepackage[most]{tcolorbox}

\definecolor{boxgold}{RGB}{252,249,240}

\title{\raisebox{-0.2cm}{\includegraphics[width=1cm]{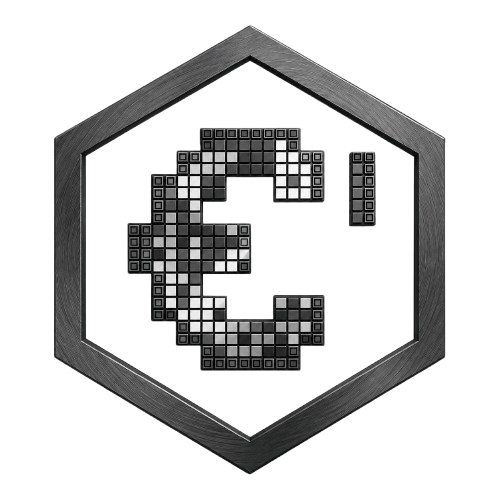}} Verify Before You Fetch: Reliable KV Cache Compression for LLM Inference}

\date{}

\author{
Devanshu Jha$^{1}$ \qquad
Gaurav Kumar$^{1}$ \qquad
Agniva Chowdhury$^{2}$
\\
Kumar Krishna Agrawal$^{3}$ \qquad
Aditya Desai$^{1}$
\\
\\
\small $^{1}$Indian Institute of Technology Bombay
\\
\small $^{2}$Rice University
\\
\small $^{3}$University of California, Berkeley
}

\begin{document}

\onecolumn
\begin{tcolorbox}[
    enhanced,
    colback=boxgold,
    colframe=boxgold,
    boxrule=0pt,
    arc=6pt,
    width=\textwidth,
    left=2pt,right=2pt,top=2pt,bottom=10pt
]
\maketitle
\begin{abstract}
\input{sections/abstract}
\end{abstract}
\end{tcolorbox}
\thispagestyle{empty}

\twocolumn

\input{sections/intro}
\input{sections/related_work.tex}
\input{sections/background}
\let\origsection\section
\renewcommand{\section}[1]{\origsection{VerKV}\let\section\origsection}
\input{sections/method_1}

\input{sections/theory_1}
\input{sections/experiments}
\input{sections/conclusion}

\bibliography{ref}
\bibliographystyle{mlsys2025}
\section*{AI Usage Disclosure}
In this work, we used AI tools to assist with polishing the writing, reviewing related literature, exploring estimator design, and refining mathematical formulations and proofs. These tools also assisted with notation consistency and LaTeX editing. We take full responsibility for the final content of this work, including all text, claims, results, and artifacts produced.

\appendix
\include{sections/appendix_theory}

\end{document}

%% file: math_commands.tex
\usepackage{amsmath}
\usepackage{amssymb}
\usepackage{amsthm}

\newtheorem{definition}{Definition}
\newtheorem{theorem}{Theorem}

\newtheorem{lemma}{Lemma}

\theoremstyle{remark}

\newcommand{\method}{\textup{\textsc{VerKV}}}
\newcommand{\kvcomp}{\textup{\textsc{PoissonComp}}}
\newcommand{\fetch}{\textup{\textsc{KVFetch}}}

%% file: sections/abstract.tex
Long-context applications, long-lived agents, and multi-agent systems require large language models (LLMs) to process and retain extensive information. In these settings, the KV cache can exceed GPU high-bandwidth memory (HBM) capacity, necessitating offloading to host memory, where CPU-to-GPU transfers can dominate inference latency. While KV cache compression can reduce data movement, query-agnostic compression that guarantees accurate attention for arbitrary queries is extremely challenging, if not impossible. We introduce \method{}, a hybrid approach that provides $(\epsilon,\delta)$ guarantees for approximating attention on any given query. \method{} maintains a compact set of query-agnostic pivotal tokens on the GPU, together with a token-level sketch of the full KV cache, while retaining the complete cache in CPU memory. We introduce a novel stochastic sampling procedure, \kvcomp{}, to select the pivotal tokens. At inference time, our proposed \fetch{} algorithm uses the pivotal tokens and the sketch to estimate the relative error of the resulting attention computation. If the estimate satisfies a user-specified $(\epsilon,\delta)$ tolerance, inference proceeds without accessing host memory. Otherwise, \method{} performs query-aware retrieval to fetch additional tokens from the CPU, refining the attention computation to meet the required tolerance. A key component of \method{} is an optimal strategy for utilizing the already sampled tokens and determining which additional tokens to fetch from the CPU, if needed. Thus, \method{} combines query-agnostic compression with query-specific verification and adaptive retrieval, enabling attention with bounded approximation error while minimizing CPU-to-GPU transfers. We show that \kvcomp{} outperforms state-of-the-art approaches to KV cache compression. Combined with \fetch{}, it achieves near-dense performance ($90$--$100\%$ of dense quality at 32K context length, averaged over RULER and LOFT), using Qwen3-4B, Qwen3.5-9B and Ministral-3-8B, while reducing the GPU memory footprint by $8$--$32\times$ and keeping CPU-to-GPU memory fetches within $2$--$23\%$ of the cache per generated token, depending on the model.

\textbf{Code:} \href{https://github.com/epsilon-prime/verifiedKV}{\mbox{https://github.com/epsilon-prime/verifiedKV}}

\textbf{Lab Page:} \href{https://epsilon-prime.github.io}{https://epsilon-prime.github.io}.%

%% file: sections/intro.tex
\section{Introduction}
\label{sec:intro}
\begin{figure*}[t]
  \centering
\includegraphics[width=0.8\textwidth]{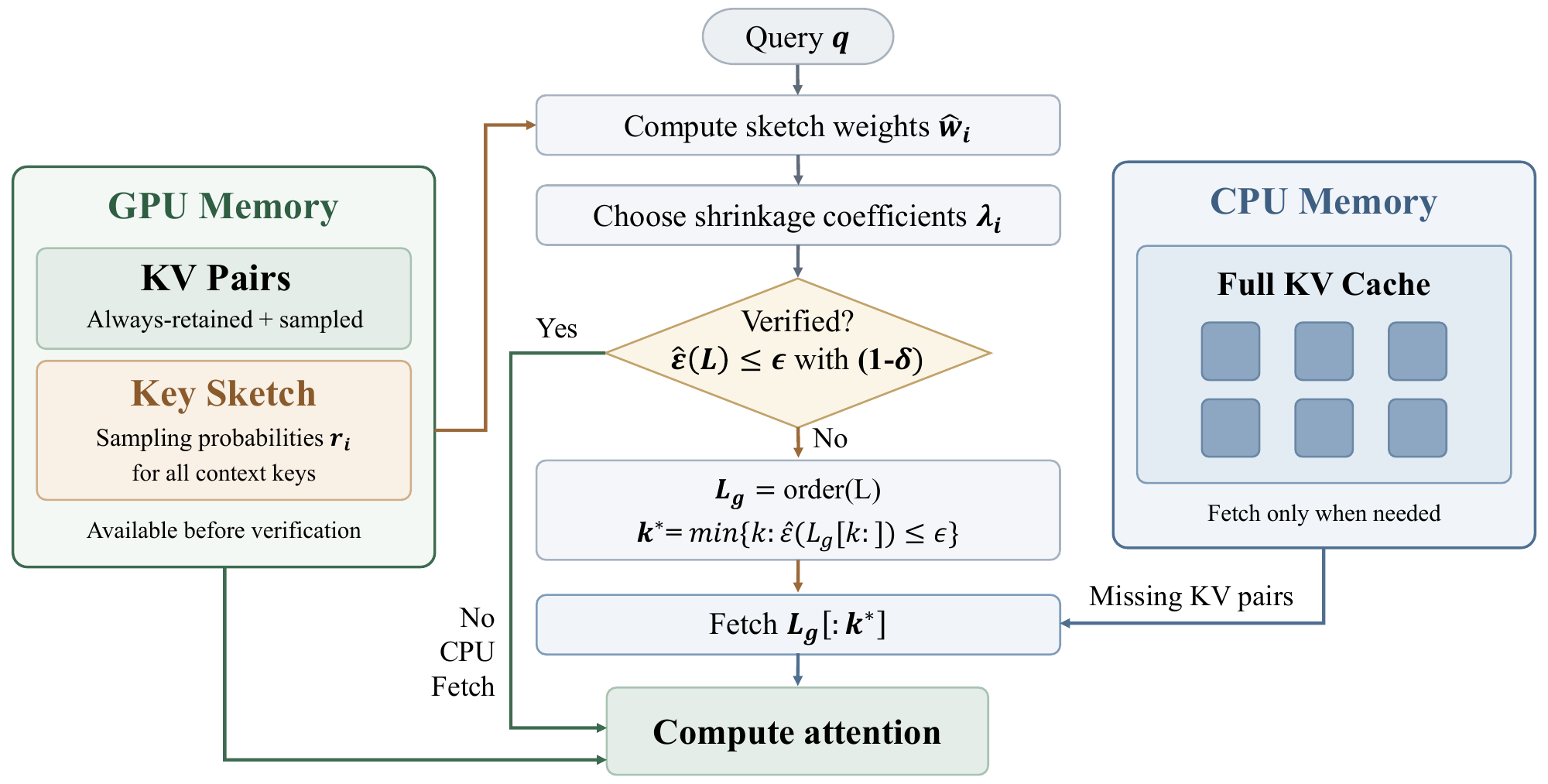}
  \caption{Overview of \method{}. The GPU holds query-agnostic pivotal tokens (deterministic and stochastic) and a sparse attention indexer (sketch), while the full KV cache resides in CPU memory. An error-estimation module uses the pivotal tokens and sketch to decide whether the approximation already meets a user tolerance $(\epsilon, \delta)$; if not, a query-aware fetch recovers additional tokens from the host so that the final attention is guaranteed to have error $(\epsilon, \delta)$.}
  \label{fig:overview}
\end{figure*}

Long context and long form generation applications~\cite{kwon2023efficient,li2023large, dosovitskiy2021image, Peebles_2023_ICCV, hou2024large, chen2024deep, Zhou2024DBGPTLL, ruan2021transformer}, long-lived agents~\cite{cao2026coding, chen2026horizon}, and multi agent systems increasingly require large language models (LLMs) to process and retain information over extended sequences. As the number of retained tokens grows, the KV cache scales linearly with sequence length, placing increasing demands on GPU memory. Despite advances in modern GPUs and their high bandwidth memory (HBM) capacity, storing the full KV cache often exceeds available memory at the context lengths required by these applications, making efficient KV cache management essential for scalable long-context inference.

Existing approaches to managing large KV caches fall into three broad categories, each with distinct limitations. KV compaction methods~\cite{cassano2026selfsummarization,openai_compaction_2026, anthropic2026compaction} compress the context in the text domain. They often use an LLM to summarize or rewrite the context and then regenerate the KV cache, incurring substantial computational overhead while potentially altering the retained information and its representation. Moreover, such methods offer no guarantees that the compressed context will preserve information relevant to future queries. KV offloading~\cite{sheng2023flexgen, lee2024infinigen} preserves the full KV cache in host memory and transfers entries to the GPU at every decoding step, avoiding approximation errors in attention computation but introducing CPU to GPU data movement that can significantly increase inference latency. KV compression~\cite{zhang2023h2o, liu2023scissorhands, li2024snapkv, kvzip, fastkvzip}, on the other hand, offers an attractive alternative by reducing the cache footprint in the embedding space without requiring context regeneration or transferring the full cache between CPU and GPU. However, despite considerable research in this direction, existing methods remain largely heuristic, providing no formal guarantees on attention approximation error for future queries and often proving brittle at high compression rates. Thus, the three approaches to managing large KV caches make fundamentally different trade-offs between future attention quality and system efficiency.

We introduce \(\method{}\), a hybrid KV cache management framework that maintains a compressed KV cache on the GPU while retaining the full cache in CPU memory. \(\method{}\) provides \((\varepsilon,\delta)\) approximation guarantees while minimizing CPU to GPU communication, eliminating transfers whenever possible. \(\method{}\) builds on two complementary approaches to different KV cache efficiency problems: query agnostic KV compression and query aware sparse attention. Existing KV compression methods assign importance scores to tokens to construct a compact representation, often by sampling from the full KV cache. This representation is then used to compute attention for all future queries, regardless of their attention patterns. In contrast, sparse attention methods use the current query and a compact sketch of the KV cache, such as a sparse indexer, to identify relevant tokens for attention computation. \(\method{}\) combines these approaches to exploit the efficiency of compressed KV representations while selectively retrieving additional KV entries from CPU memory when needed to satisfy the desired approximation guarantees.

First, \(\method{}\) uses the importance scores of an existing KV compression method to parameterize Poisson sampling, producing a stochastic sample of pivotal tokens that remains resident on the GPU. Unlike deterministic selection, this sampling enables estimation of both the attention output and its approximation error. Second, \(\method{}\) leverages the compact sketches used by sparse attention methods to compute proxy scores for \(\langle k_i,q\rangle\) for the current query. Combined with the sampling probabilities, these scores help refine the error estimates and identify additional tokens to fetch from CPU memory when needed.

Figure~\ref{fig:overview} illustrates the overall design. After prefill, \(\method{}\) constructs the GPU resident sample and its associated sketch. At each decoding step, it first estimates the approximation error of attention computed using only the sampled tokens and the GPU-resident sketch. If the error is bounded by \(\varepsilon\) with probability at least \(1-\delta\), it avoids CPU fetches entirely. Otherwise, it selectively fetches additional tokens using the sketch and sampling probabilities, minimizing data movement while ensuring that the resulting attention approximation satisfies the target error bound.

We evaluate \method{} on three models (Qwen3-4B, Qwen3.5-9B, Ministral-3-8B) and two long-context benchmarks (RULER and LOFT) at 32K and 128K context. At matched GPU memory, query-agnostic eviction collapses as compression grows: Fast KVzip and KVzip retain only $2$--$12\%$ of dense quality at extreme compression. \method{} achieves near-dense performance over both the 32K and 128K datasets: on average, it retains $96\%$ of dense quality at $8\times$, going to $86\%$ at $32\times$ (Table~\ref{tab:punchline}). \method{} achieves this while fetching only $2$--$23\%$ of the KV cache per generated token. Sampling alone (\kvcomp{}) already beats top-$k$ eviction, and the certificate-driven fetch further recovers majority of the remaining loss. With a perfect sketch, the per-decision error stays within $\varepsilon_0$ in $99.3$--$99.8\%$ of decisions, well inside $\delta$.
Implemented in vLLM, \method{} achieves $1.6\times$ the decode throughput of offloaded dense attention, at near-dense quality, all while keeping only $6.25\%$ of the KV cache resident on the GPU (Section~\ref{sec:systems-pareto}).

\begin{table*}[t]
  \centering
\caption{Quality, GPU-resident memory and host traffic of dense attention, KVzip, Fast KVzip and \method{} on three models. Quality is the mean fraction of dense performance across 18 tasks spanning RULER (multi-key and multi-value needle retrieval, common-word extraction, question answering) and LOFT (NQ, HotpotQA, MuSiQue, QAMPARI, QUEST) at 32K and 128K context ($\mathcal{C}=$ Fast KVzip, $\mathcal{A}=$ 2-bit sketch, $\varepsilon_0=0.05$, $\delta=0.1$). \method{} delivers a score comparable to dense attention, while keeping a low GPU-resident memory and simultaneously fetching a fraction of the full cache per generated token.}
  \label{tab:punchline}
  \small
  \input{image_assets/punchline_table/punchline_table}
\end{table*}

%% file: image_assets/punchline_table/punchline_table.tex
\setlength{\tabcolsep}{4pt}
\begin{tabular}{l|rrr|rrr|rrr}
\toprule
 & \multicolumn{3}{c|}{\textbf{Qwen3-4B}} & \multicolumn{3}{c|}{\textbf{Qwen3.5-9B}} & \multicolumn{3}{c}{\textbf{Ministral-3-8B}} \\
 & Quality \% & \% Memory & \% Fetch & Quality \% & \% Memory & \% Fetch & Quality \% & \% Memory & \% Fetch \\
 &  & resident & /token &  & resident & /token &  & resident & /token \\
\midrule
Dense (resident) & \cellcolor[rgb]{0.700,0.870,0.660}100 & \cellcolor[rgb]{0.930,0.360,0.330}100 & \cellcolor[rgb]{0.700,0.870,0.660}0 & \cellcolor[rgb]{0.700,0.870,0.660}100 & \cellcolor[rgb]{0.930,0.360,0.330}100 & \cellcolor[rgb]{0.700,0.870,0.660}0 & \cellcolor[rgb]{0.700,0.870,0.660}100 & \cellcolor[rgb]{0.930,0.360,0.330}100 & \cellcolor[rgb]{0.700,0.870,0.660}0 \\
Dense (offload) & \cellcolor[rgb]{0.700,0.870,0.660}100 & \cellcolor[rgb]{0.700,0.870,0.660}0 & \cellcolor[rgb]{0.930,0.360,0.330}100 & \cellcolor[rgb]{0.700,0.870,0.660}100 & \cellcolor[rgb]{0.700,0.870,0.660}0 & \cellcolor[rgb]{0.930,0.360,0.330}100 & \cellcolor[rgb]{0.700,0.870,0.660}100 & \cellcolor[rgb]{0.700,0.870,0.660}0 & \cellcolor[rgb]{0.930,0.360,0.330}100 \\
\midrule
KVzip $8\times$ & \cellcolor[rgb]{0.892,0.897,0.607}66.9 & \cellcolor[rgb]{0.772,0.880,0.640}12.5 & \cellcolor[rgb]{0.700,0.870,0.660}0 & \cellcolor[rgb]{0.950,0.547,0.415}17.0 & \cellcolor[rgb]{0.772,0.880,0.640}12.5 & \cellcolor[rgb]{0.700,0.870,0.660}0 & \cellcolor[rgb]{0.966,0.907,0.587}54.1 & \cellcolor[rgb]{0.772,0.880,0.640}12.5 & \cellcolor[rgb]{0.700,0.870,0.660}0 \\
KVzip $16\times$ & \cellcolor[rgb]{0.953,0.571,0.426}19.2 & \cellcolor[rgb]{0.736,0.875,0.650}6.25 & \cellcolor[rgb]{0.700,0.870,0.660}0 & \cellcolor[rgb]{0.933,0.388,0.343}2.6 & \cellcolor[rgb]{0.736,0.875,0.650}6.25 & \cellcolor[rgb]{0.700,0.870,0.660}0 & \cellcolor[rgb]{0.965,0.677,0.474}28.8 & \cellcolor[rgb]{0.736,0.875,0.650}6.25 & \cellcolor[rgb]{0.700,0.870,0.660}0 \\
KVzip $32\times$ & \cellcolor[rgb]{0.940,0.447,0.370}8.0 & \cellcolor[rgb]{0.718,0.873,0.655}3.13 & \cellcolor[rgb]{0.700,0.870,0.660}0 & \cellcolor[rgb]{0.932,0.381,0.340}1.9 & \cellcolor[rgb]{0.718,0.873,0.655}3.13 & \cellcolor[rgb]{0.700,0.870,0.660}0 & \cellcolor[rgb]{0.937,0.422,0.358}5.6 & \cellcolor[rgb]{0.718,0.873,0.655}3.13 & \cellcolor[rgb]{0.700,0.870,0.660}0 \\
\midrule
Fast KVzip $8\times$ & \cellcolor[rgb]{0.903,0.898,0.604}65.0 & \cellcolor[rgb]{0.772,0.880,0.640}12.5 & \cellcolor[rgb]{0.700,0.870,0.660}0 & \cellcolor[rgb]{0.950,0.545,0.414}16.8 & \cellcolor[rgb]{0.772,0.880,0.640}12.5 & \cellcolor[rgb]{0.700,0.870,0.660}0 & \cellcolor[rgb]{0.907,0.899,0.603}64.3 & \cellcolor[rgb]{0.772,0.880,0.640}12.5 & \cellcolor[rgb]{0.700,0.870,0.660}0 \\
Fast KVzip $16\times$ & \cellcolor[rgb]{0.964,0.673,0.472}28.4 & \cellcolor[rgb]{0.736,0.875,0.650}6.25 & \cellcolor[rgb]{0.700,0.870,0.660}0 & \cellcolor[rgb]{0.943,0.477,0.383}10.6 & \cellcolor[rgb]{0.736,0.875,0.650}6.25 & \cellcolor[rgb]{0.700,0.870,0.660}0 & \cellcolor[rgb]{0.973,0.758,0.511}36.2 & \cellcolor[rgb]{0.736,0.875,0.650}6.25 & \cellcolor[rgb]{0.700,0.870,0.660}0 \\
Fast KVzip $32\times$ & \cellcolor[rgb]{0.940,0.448,0.370}8.0 & \cellcolor[rgb]{0.718,0.873,0.655}3.13 & \cellcolor[rgb]{0.700,0.870,0.660}0 & \cellcolor[rgb]{0.943,0.478,0.384}10.7 & \cellcolor[rgb]{0.718,0.873,0.655}3.13 & \cellcolor[rgb]{0.700,0.870,0.660}0 & \cellcolor[rgb]{0.944,0.491,0.389}11.9 & \cellcolor[rgb]{0.718,0.873,0.655}3.13 & \cellcolor[rgb]{0.700,0.870,0.660}0 \\
\midrule
\method{} $8\times$ & \cellcolor[rgb]{0.700,0.870,0.660}100.1 & \cellcolor[rgb]{0.772,0.880,0.640}12.5 & \cellcolor[rgb]{0.715,0.872,0.656}2.6 & \cellcolor[rgb]{0.755,0.878,0.645}90.6 & \cellcolor[rgb]{0.772,0.880,0.640}12.5 & \cellcolor[rgb]{0.712,0.872,0.657}2.1 & \cellcolor[rgb]{0.710,0.871,0.657}98.2 & \cellcolor[rgb]{0.772,0.880,0.640}12.5 & \cellcolor[rgb]{0.756,0.878,0.645}9.6 \\
\method{} $16\times$ & \cellcolor[rgb]{0.729,0.874,0.652}95.0 & \cellcolor[rgb]{0.736,0.875,0.650}6.25 & \cellcolor[rgb]{0.726,0.874,0.653}4.4 & \cellcolor[rgb]{0.786,0.882,0.636}85.2 & \cellcolor[rgb]{0.736,0.875,0.650}6.25 & \cellcolor[rgb]{0.719,0.873,0.655}3.3 & \cellcolor[rgb]{0.786,0.882,0.636}85.2 & \cellcolor[rgb]{0.736,0.875,0.650}6.25 & \cellcolor[rgb]{0.783,0.881,0.637}14.3 \\
\method{} $32\times$ & \cellcolor[rgb]{0.731,0.874,0.651}94.6 & \cellcolor[rgb]{0.718,0.873,0.655}3.13 & \cellcolor[rgb]{0.749,0.877,0.646}8.5 & \cellcolor[rgb]{0.789,0.882,0.635}84.6 & \cellcolor[rgb]{0.718,0.873,0.655}3.13 & \cellcolor[rgb]{0.737,0.875,0.650}6.3 & \cellcolor[rgb]{0.818,0.886,0.627}79.6 & \cellcolor[rgb]{0.718,0.873,0.655}3.13 & \cellcolor[rgb]{0.831,0.888,0.624}22.6 \\
\bottomrule
\end{tabular}

%% file: sections/related_work.tex
\section{Related Work}
\label{sec:related-work}
We organize the closest work into three complementary
categories: KV compaction, which shortens the context;
KV compression, which reduces cache storage; and sparse
attention, which selects the entries accessed for each query.

\textbf{KV compaction.}
Context compaction replaces earlier interactions with
a shorter representation, often a summary.
Claude~\cite{anthropic2026compaction} provides
summarization-based compaction, while
Composer~\cite{cassano2026selfsummarization} incorporates
self-summarization into training for long-horizon tasks.
In \method{}, the original KV cache remains in CPU
memory, allowing entries absent from the GPU to be
accessed again.

\textbf{KV compression.}
KV compression operates directly on cached keys and
values without rewriting the input context.
Token eviction methods include
H$_2$O~\cite{zhang2023h2o},
Scissorhands~\cite{liu2023scissorhands}, and
SnapKV~\cite{li2024snapkv}.
KVzip~\cite{kvzip} estimates token importance through
context reconstruction, while Fast KVzip~\cite{fastkvzip}
uses lightweight gating to reduce the cost of importance
estimation.
Compactor~\cite{chari2025compactor} uses approximate
leverage scores for query-agnostic compression.
\method{} uses such estimates to set Bernoulli inclusion
probabilities for GPU-resident entries.

\looseness=-1\textbf{Sparse attention.}
Sparse attention selects entries at decode time using
the current query.
Quest~\cite{tang2024quest},
HashAttention~\cite{desai2025hashattention}, and
SOCKET~\cite{joshi2026socket} use compact representations
to identify relevant entries without scoring every full key.
Sampling offers another approach:
MagicPIG~\cite{chen2024magicpig} uses locality-sensitive
hashing to sample attention contributions, while
vAttention~\cite{desai2026vattention} combines top-$k$
selection with sampling to control approximation error. In contrast,
\method{} focuses on verification with a GPU-resident
sample selected during prefill.
It chooses query-specific shrinkage coefficients and
uses a denominator certificate to determine which
additional contributions to make exact, fetching their
KV pairs only when they are absent from the GPU
(Section~\ref{sec:method}).

%% file: sections/background.tex
\section{Background}

For keys and values $\mathbf K,\mathbf V\in\mathbb R^{n\times d}$
and a query $q\in\mathbb R^d$, define
$w_i=\exp(\langle\mathbf K[i],q\rangle/\sqrt d)$.
Full attention computes
\begin{equation}
o=A(\mathbf K,\mathbf V,q)
=\frac{N}{D}
=\frac{\sum_{i=0}^{n-1}w_i\mathbf V[i]}
       {\sum_{i=0}^{n-1}w_i},
\end{equation}
where $N$ and $D$ denote the numerator and denominator.
Given a reduced cache
$\widehat{\mathbf K},\widehat{\mathbf V}\in\mathbb R^{m\times d}$
with $m<n$, the approximate output is
$\widehat o=A(\widehat{\mathbf K},\widehat{\mathbf V},q)$.
We measure its error by $\|\widehat o-o\|_2$.
The following formulations differ in when entries are
selected and whether the current query is available.

\looseness=-1\textbf{KV Cache Compression. (Algorithm  $\mathcal{C}$)}
KV cache compression selects a subset of KV pairs during prefill, before future query vectors are available. A general approach uses a scoring function \(\mathcal{S}(\mathbf{K}, \mathbf{V}; \mathbf{Q})\) to estimate the importance of each KV pair based on the query vectors \(\mathbf{Q}\) observed while processing the current input, and retains the top \(m\) pairs.
We choose KVzip~\cite{kvzip} and Fast KVzip~\cite{fastkvzip} for experiments. We also implement Fast KVzip in vLLM for system measurements.

\textbf{Sparse Attention (Decode). (Algorithm $\mathcal{A}$)}
Unlike KV cache compression, sparse attention selects a subset of KV pairs at decode time, once the current query \(q\) is available. It requires the entire KV cache to remain accessible and uses a query-aware scoring function \(\mathcal{S}(\mathbf{K}, \mathbf{V}; q)\) to estimate the importance of each KV pair and select the top \(m\) pairs for attention computation. To score each KV token, it uses a compact representation of the KV cache, say $\mathcal{M}$, which gives us a way to estimate $\langle q, \mathbf{K}[i] \rangle$ for all tokens $i$.
Section~\ref{sec:related-work} reviews related methods.

The above formulations capture a broad class of methods for these two problems. However, other approaches, including stochastic sampling and learned representations, use different mechanisms to approximate attention and are not directly captured by these formulations.

%% file: sections/method_1.tex
\section{VerKV}
\label{sec:method}
We borrow the definition of verified computation under approximation from~\citet{desai2026vattention}. The definition is as follows,

\begin{definition}[$(\epsilon,\delta)$-verified-$\mathcal{X}$]
For any given computation $\mathcal{X} : \mathbb{R}^{d_1} \to \mathbb{R}^{d_2}$ for some $d_1,d_2 \in \mathbb{N}$, an algorithm $\mathcal{X}'$ is $(\epsilon,\delta)$-verified-$\mathcal{X}$ if the following holds for any $x \in \mathbb{R}^{d_1}$:
\begin{equation}
    \Pr\left(
        \frac{\|\mathcal{X}'(x)-\mathcal{X}(x)\|_2}
        {\|\mathcal{X}(x)\|_2}
        > \epsilon
    \right) < \delta.
\end{equation}
\end{definition}
Ideally, we would like to verify the approximation with respect to \(\mathcal{X}\), representing the entire model. However, analyzing such a functional object is extremely challenging. We therefore focus on smaller computations that encapsulate the approximation error. For KV cache approximation, a natural choice is to consider the numerator and denominator of the attention computation, denoted by \(\mathcal{N}\) and \(\mathcal{D}\), respectively. Given an \((\epsilon_1,\delta_1)\)-verified-\(\mathcal{D}\) algorithm and an \((\epsilon_2,\delta_2)\)-verified-\(\mathcal{N}\) algorithm, we can combine them to obtain an \((\epsilon,\delta)\)-verified SDPA algorithm (Lemma 4.2 of \cite{desai2026vattention}). Since both the numerator and denominator are sums of \(n\) terms, the mathematical machinery used to analyze them is largely similar. In what follows, we present a \((\epsilon,\delta)\)-verified-\(\mathcal{D}\) algorithm, which is also the approach we implement in our experiments.

\textbf{Overview.} An overview of our algorithm is shown in Figure~\ref{fig:overview}. \method{} is parameterized by a KV compression algorithm \(\mathcal{C}\) and a sparse attention algorithm \(\mathcal{A}\). We introduce \kvcomp{}, a new KV cache compression algorithm that uses Poisson sampling based on the scores \(\mathcal{S}_\mathcal{C}(\mathbf{K}, \mathbf{V}; \mathbf{Q})\) provided by \(\mathcal{C}\). \kvcomp{} itself provides a Pareto improvement over existing KV compression approaches. The compressed KV cache, \(\hat{\mathbf{K}}, \hat{\mathbf{V}}\), is typically generated after prefilling a long context and stored in GPU memory. Depending on the application, compression may be repeated intermittently throughout the request lifecycle. We also retain a small number of sink and local tokens throughout the request.

At each decoding step, we invoke \fetch{}, which is central to our \((\epsilon,\delta)\)-verified-\(\mathcal{D}\) recipe. First, \fetch{} determines whether \(\mathcal{D}\) can be computed using only \(\hat{\mathbf{K}}, \hat{\mathbf{V}}\) within the specified \((\epsilon,\delta)\) error bounds. If so, it completely avoids CPU fetches. Otherwise, \fetch{} identifies the minimum set of tokens that must be fetched from CPU memory to ensure that the approximation satisfies the same error bounds. Thus, at every decoding step, we obtain an \((\epsilon,\delta)\)-verified computation of \(\mathcal{D}\). In the remainder of this section, we describe the \kvcomp{} and \fetch{} algorithms in detail.
\begin{algorithm}[t]
  \caption{\method{}:\kvcomp{}}
  \label{alg:prefill}
\begin{algorithmic}[1]
  \REQUIRE context of $n$ tokens; kept fraction $\eta$, KV Compression algorithm $\mathcal{C}$
  \STATE $\tau = \mathcal{S}(\mathbf{K}, \mathbf{V}; \mathbf{Q})$, $\tau[i] = 1 $ if $i$ in sink / local windows
  \STATE find $c$ s.t $\sum_i\min(1,c\tau_i)=\eta n$
  \STATE $r_i\gets\min(1,c\tau_i)$,$H\gets\{i:r_i{=}1\}$, $L\gets\{i:r_i{<}1\}$
  \STATE $Z_i\sim_{independent}\mathrm{Bernoulli}(r_i)$  $\forall i\in L$
    \STATE keep $H\cup\{i\in L:Z_i=1\}$ on the GPU \\
    \COMMENT{After compression, rest of the KV cache is offloaded and we store all $r$ on GPU for future computation.}
\end{algorithmic}
\end{algorithm}

\subsection{\texorpdfstring{\kvcomp{}($\mathcal{C}, \mathcal{A}, \eta$)}{PoissonComp(C, A, eta)}}
\kvcomp{} always retains a few sink and local tokens. For the remaining tokens, given a budget of \(\eta\) tokens, \kvcomp{} uses the importance scores \(\tau = \mathcal{S}_\mathcal{C}(\mathbf{K}, \mathbf{V}; \mathbf{Q})\) provided by the KV compression algorithm \(\mathcal{C}\) parameterizing \kvcomp{} to define the inclusion probabilities:
\begin{equation}
    r_i = \min\{1, c \tau_i\}.
\end{equation}
$c$ is chosen such that $\sum_i \min(1, c\tau_i) = \eta n$
We assume without loss of generality that \(\tau_i \in [0,1]\). If the scores do not lie in this range, we can transform them using a softmax. Scaling the scores by $c$ ensures that the expected sample size is proportional to the specified budget. This formulation also ensures that high-scoring tokens are always included. Let $H$ be the set of all tokens retained with probability $1$ including sink and local tokens and \(L = \{i : 0 < r_i < 1\}\). We then construct a Poisson sample:
\[
S = \{i \in L : Z_i = 1\}, \qquad Z_i \sim \mathrm{Bernoulli}(r_i),
\]
where all \(Z_i\) are independent. We retain sample $S$ along with $H$ on the GPU.
\begin{algorithm}[t]
  \caption{\method{}: \fetch{}}
  \label{alg:vkv-decision}
  \begin{algorithmic}[1]
    \REQUIRE $\mathcal{M}_{\mathcal{A}}$ Sketch corresponding to algorithm $\mathcal{A}$, $L$ and $H$ index sets, $r$ inclusion probabilities. Sampled set $S$, tolerance  $(\epsilon, \delta)$
    \STATE Compute $\hat{w}_i \forall i$ using $\mathcal{M}_\mathcal{A}$
    \STATE $\lambda \gets$ bisection minimizing MSE \eqref{eq:mse}
    \IF{$\widehat\varepsilon(L)<\epsilon$
     (see~\eqref{eq:epsiloncomputation})}
        \STATE Return $H,L$
    \ENDIF
    \STATE Compute $g$, s.t, $g_i = \widehat\varepsilon(L) - \widehat\varepsilon(L \setminus \{i\})$
    \STATE $L_g = sort(L)$ descending by $g$
    \STATE Choose $k^\star=\min\{k\in\{0,\ldots,|L|\}:
\widehat\varepsilon(L_g[k:])<\epsilon\}$.
    \STATE Set $L'=L_g[k^\star:]$ and fetch
$(L\setminus L')\setminus S$.
    \STATE Set $H'=H \cup L_g[:k^\star]$.
    \STATE Return $H',L'$ \\
    \COMMENT{ Compute attention using ordinary weights on $H'$ and weights $\lambda_i \hat{w}_i/r_i$ on $S\cap L'$.}
  \end{algorithmic}
\end{algorithm}

\subsection{\fetch{}($\mathcal{C}, \mathcal{A}, \epsilon, \delta$)}
At any given decode step with query $q$, we have $H$ (indices kept with probability $1$) and $L$ (indices kept with probability $< 1$ ). The \fetch{} algorithm chooses which tokens to fetch from CPU, i.e., which tokens to move from set $L$ to set $H$. We want to make the minimal movement needed to ensure the $(\epsilon, \delta)$ guarantee. We will first explain which estimator to use given a specific $H$ and $L$. Then, we will describe how to make minimal fetches to update $H$ in a way that preserves the guarantee.
Our analysis considers a fixed query $q$ that is independent of the initial Poisson sample. All probabilities and expectations are taken over the randomness of this sample. For analytical simplicity, we assume perfect sketch estimation, i.e., $\widehat w_i = w_i$ for all $i \in L$, and use $w_i$ throughout the proofs. The effect of deviations of $\widehat w_i$ from $w_i$ can also be analyzed but is left for future work. In practice, all implementations use $\widehat w_i$. In our experiments, we report results with both perfect and approximate sketches. The effect of sketch approximation on the error guarantee is shown in Figure~\ref{fig:adherence}.

Now, recall that we are estimating,
\begin{equation}
    D = \sum_{i \in H} w_i  + \sum_{i \in L} w_i
\end{equation}

For the given sets $H$ and $L$, we use the following
estimator.

\begin{definition}[Bounded Shrinkage Estimator] Given a query $q$, given set of indices $H$ and $L$ with probabilities of inclusion $\mathbf{r}$ for L, the bounded shrinkage estimator used in \kvcomp{} is
\begin{equation}
    \hat{D} = \sum_{i \in H} w_i + \sum_{i \in L} Z_i \lambda_i \frac{w_i}{r_i}
\end{equation}
where
$Z_i \sim \operatorname{Bernoulli}(r_i)$, $\lambda_i
=
\operatorname{clip}
_{[r_i,1]}
\left(
\frac{\beta r_i}
{w_i(1-r_i)}
\right)$,
with
$\operatorname
{clip}
_{[r_i,1]}
(x)
=\min\{1,\max\{r_i,x\}\}$ and
$\beta
=
\sum_{i\in L} w_i(1-\lambda_i)$.
\label{def:estimator}
\end{definition}

If we choose $\lambda_i = 1$, we obtain the usual unbiased estimator. However, we can obtain a better estimator by trading off variance for bias. For instance, when $r_i$ is small, the unbiased contribution of a token can introduce high variance. Thus, instead of requiring an unbiased estimator, we aim to minimize the mean squared error (MSE). We can show that,
\begin{theorem}
The bounded shrinkage estimator defined in Definition~\ref{def:estimator} uniquely minimizes the Mean Square Error
\begin{equation}
\label{eq:mse}
\operatorname{MSE}_{\widehat D}(\lambda)
=
\left(\sum_{i\in L}w_i(1-\lambda_i)\right)^2
+
\sum_{i\in L}\lambda_i^2w_i^2\frac{1-r_i}{r_i}.
\end{equation}
over $\lambda$ such that $\lambda_i \in [r_i, 1] \forall i$.
\end{theorem}

\begin{figure}[t]
  \centering
  \includegraphics[width=\columnwidth]{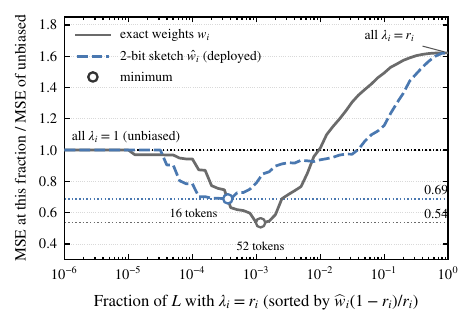}
  \caption{Binary shrinkage. In practice we apply a simpler bound, with $\lambda_i\in\{1,r_i\}$. The figure shows the MSE of the denominator estimate for one query row (Qwen3-4B, RULER-32K, $16\times$) as more tokens of $L$ are switched from $\lambda_i=1$ to $\lambda_i=r_i$, relative to the unbiased estimator. Shrinking a few tokens cuts the MSE nearly in half, while shrinking all of them makes it worse ($1.62\times$); the deployed sketch finds a slightly earlier, shallower minimum than exact weights. See Appendix~\ref{app:binary-shrinkage}.}
  \label{fig:binary-shrinkage-main}
\end{figure}

The motivation for using bounded shrinkage is to avoid overreliance on the quality of the estimates $\hat{w}_i$. While these estimates are useful for preventing obvious increases in variance, they may not be sufficiently accurate to support unbounded shrinkage. In practice, we can use different kinds of bounds on $\lambda$, such as using discrete $\lambda_i\in\{1,r_i\}$. The impact of shrinkage on MSE is shown in Figure~\ref{fig:binary-shrinkage-main}).

We now establish the Bernstein~\cite{bernstein1924modification} certificate for
Definition~\ref{def:estimator}. Note that the same proof applies
to any fixed coefficients $\lambda_i\in[r_i,1]$
chosen independently of the sampling outcomes,
including those selected from $\lambda_i\in\{r_i,1\}$

\begin{lemma}\label{lemma:bernstien}
For $0<\delta<1$, the estimator $\widehat D$ in
Definition~\ref{def:estimator} satisfies
$|\widehat D-D|/D\le\widehat\varepsilon(L)$
with probability at least $1-\delta$, where
\begin{equation}
\label{eq:epsiloncomputation}
\widehat\varepsilon(L)
=
\frac{
\beta(L)+\frac{b(L)l_\delta}{3}
+
\sqrt{
\left(\frac{b(L)l_\delta}{3}\right)^2
+2V(L)l_\delta
}
}{D}.
\end{equation}
Here $l_\delta=\log(2/\delta)$ and $D$ is the full
attention denominator, $\beta(L)=
\sum_{i\in L} \left( w_i - \lambda_i w_i \right)$, $V(L)=
\sum_{i\in L} \left( \lambda_i^2 w_i^2 \frac{1 - r_i }{r_i} \right)$, and $b(L)= \max_{i\in L} \left( \frac{\lambda_i w_i}{r_i} \max(r_i , 1 - r_i) \right)$. Empty sums and maxima are defined as zero.
\end{lemma}
{The certificate combines three quantities computed over the tokens still in $L$.
$\beta(L)$ is the bias from shrinkage: the total weight the estimator gives up by using $\lambda_i<1$, which is zero for the unbiased estimator ($\lambda_i=1$).
$V(L)$ is the variance from sampling, dominated by tokens with a large weight and a small inclusion probability.
$b(L)$ is the largest amount any single token can move the estimate away from its mean.
Bernstein's inequality turns $V(L)$ and $b(L)$ into a deviation bound that holds with probability at least $1-\delta$; adding the bias and dividing by $D$ gives the relative error $\widehat\varepsilon(L)$.
Moving a token from $L$ to $H$ removes its contribution from all three quantities, so each fetched token can only tighten the bound.}

\textbf{Case: No Fetch.} The above immediately gives us the capability to decide whether we need to fetch anything from CPU. At any decode step, if $\widehat\varepsilon(L) < \epsilon$, then we can use the estimator $\hat{D}$ to estimate using GPU resident $\hat{K}, \hat{V}$ given by \kvcomp{}.

\textbf{Case: Fetch.} In case, $\widehat\varepsilon(L) > \epsilon$, \fetch{} algorithm needs to find a subset $L' \subset L$ of maximum cardinality such that $\varepsilon(L') < \epsilon$.
\begin{equation}
L' = \arg_{L' \subseteq L} \max_{ \widehat\varepsilon(L') < \epsilon} |L'| \label{eq:fetchmax}
\end{equation}
Once such a set is found, \fetch{} fetches $L - L'$ and updates $H' = H \cup (L-L')$. Then we can use the estimator $\hat{D}(H',L')$

Solving the eq.~\eqref{eq:fetchmax} exactly is nontrivial given constraints of efficiency. Instead, we use a greedy algorithm to find a set $L'$ which satisfies $\widehat\varepsilon(L') < \epsilon$. We describe the algorithm below.

We first order indices in $L$ in descending order of importance score $g_i$ where we measure importance by the drop in $\varepsilon$ if we were to fetch this token.
\begin{equation}
\label{eq:fetch-score}
g_i
=   \widehat\varepsilon(L) - \widehat\varepsilon(L \setminus \{i\})
 \end{equation}

where $\widehat\varepsilon(L)$ is defined in Equation~\eqref{eq:epsiloncomputation}.
{Let $L_g$ denote $L$ sorted in descending order of $g_i$,}  Then we want to find a minimum $k$ such that $\widehat\varepsilon(L[k:]) < \epsilon$
\begin{equation}
k^\star = \min \left\{ k \in \{0,\ldots, |L|\} : \widehat\varepsilon(L_g[k:]) < \epsilon \right\}
\end{equation}
We make these $k^\star$ contributions exact,
fetching only those not already in $S$.

\begin{theorem}
\label{thm:verified-denominator}
Under the fixed-query, exact-sketch, and Poisson
sampling assumptions of
Lemma~\ref{lemma:bernstien}, let $\epsilon>0$
and $0<\delta<1$.
Suppose the coefficients $\lambda_i\in[r_i,1]$
and the ordering $L_g$ are chosen independently
of the sampling outcomes, and the coefficients
are held fixed throughout fetching. For each $k$, evaluate the certificate on the
residual set $L_g[k:]$, using these fixed
coefficients and the original denominator
$D$.
Then
\[
k^\star
=
\min\left\{
k\in\{0,\ldots,|L|\}:
\widehat\varepsilon(L_g[k:])<\epsilon
\right\}
\]
exists.
Writing $L'=L_g[k^\star:]$, the returned estimator
\[
\widehat D
=
\sum_{i\in H}w_i
+
\sum_{i\in L\setminus L'}w_i
+
\sum_{i\in L'}\frac{Z_i\lambda_iw_i}{r_i}
\]
satisfies
\[
\Pr\!\left(
\frac{|\widehat D-D|}{D}>\epsilon
\right)
\le\delta.
\]
\end{theorem}

%% file: sections/experiments.tex
\section{Experiments}
\label{sec:experiments}

We evaluate \method{} across three models, two long-context benchmarks, two context lengths and three compression rates, and measure its speed in an implementation inside vLLM. Our key findings are the following.

\begin{figure*}[!t]
  \centering
  \includegraphics[width=\textwidth]{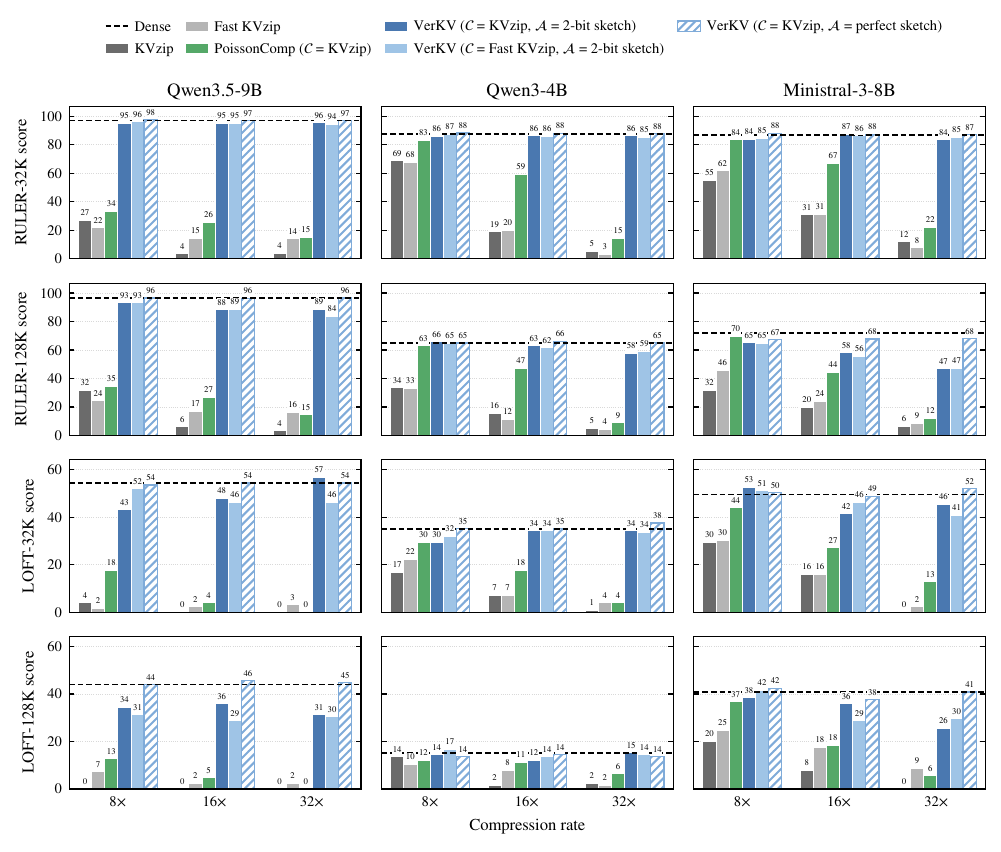}  %
\caption{Quality vs.\ compression rate. Average quality of different compression setups across the RULER and LOFT benchmarks at context lengths of 32K and 128K for three models. We observe that (1) \kvcomp{}($\mathcal{C}$=KVzip) {consistently outperforms KVzip}, and (2) \method{}($\mathcal{C}$=KVzip, $\mathcal{A}$=2-bit sketch) = \kvcomp{}($\mathcal{C}$=KVzip) + \fetch{} improves upon \kvcomp{} under all settings. (3) Relative to dense attention, quality is better preserved at shorter context lengths for a given compression rate, for the Qwen family compared to Ministral, and on RULER compared to LOFT. In particular, \method{} recovers near-dense quality on RULER-32K across all three models. Our algorithm currently uses a naive 2-bit sketch to bootstrap its estimates; improving sketch accuracy is a promising direction for further quality gains as can be seen with \method{}($\mathcal{C}$=KVzip, $\mathcal{A}$=perfect sketch) which matches Dense in nearly every setting. Host traffic for these runs is reported per generated token in Table~\ref{tab:punchline} and per forward pass in Table~\ref{tab:fetch-stats}.}
\label{fig:quality-compression}
\end{figure*}

\begin{itemize}[leftmargin=*, nosep]
\item \kvcomp{} consistently outperforms deterministic KV compression methods across models, context lengths, and datasets. Combining \kvcomp{} with selective fetching \fetch{} significantly closes the quality gap with dense attention, with the extent of improvement depending on the quality of the sketch. With a perfect sketch, we completely close the gap across all settings. Even with a naive 2-bit sketch, representative of sparse attention indexers, fetching substantially narrows the gap while still falling short of dense attention. (Figure~\ref{fig:quality-compression}). Notably a good sketch can nearly eliminate the gap to dense with $32\times$ less GPU memory usage.

 \item \fetch{} can often avoid CPU access entirely and, when fetching is necessary, retrieves only a fraction of the KV cache. These rates are effectively controlled through the externally specified tolerance parameters $(\epsilon, \delta)$. At {$\varepsilon_0=0.05$ }, \fetch{} is able to compute attention using only GPU resident memory for about {$20$ -- $25\%$} of the computations. Moreover, it only {reads $9$--$30\%$ of the KV cache per forward pass across models}, substantially reducing data movement compared to full dense offloading. (Table~\ref{tab:fetch-stats})

 \item The selective fetching also helps us improve throughput over full offloading dense computation. On Qwen3-4B in our vLLM fork, \method{} decodes $1.6\times$ faster than offloading the full cache because it transfers $47\times$ less data from host memory, while staying close to the quality of dense model ( $86.7$ against $88.0$ for Dense) (Section~\ref{sec:systems-pareto}).
 \end{itemize}

\subsection{Methodology}
\label{sec:setup}

\looseness=-1\textbf{Implementation.}
We measure quality with a reference implementation of \method{} and all baselines in the KVPress evaluation pipeline~\citep{kvpress2024}, and speed with an implementation of \method{} inside vLLM~\citep{kwon2023efficient}, described in Appendix~\ref{app:vllm}.
The vLLM experiments run on one NVIDIA H100 GPU and serve one request at a time, except for the batch-size study in Appendix~\ref{app:vllm}.

\looseness=-1\textbf{Models and benchmarks.}
We evaluate three instruction-tuned models from different families, all in BF16: Qwen3-4B-Instruct-2507~\citep{yang2025qwen3}, Qwen3.5-9B~\citep{qwen35blog} and Ministral-3-8B-Instruct-2512~\citep{liu2026ministral}.
We use RULER~\citep{hsieh2024ruler} and LOFT~\citep{Lee2024Loft}, each at $32$K and $128$K context.
From RULER we take four tasks that cover needle retrieval, and question answering: multi-key needle retrieval (MK3), multi-value needle retrieval (MV), common-word extraction (CWE) and single-hop question answering (QA1).
Each task uses $50$ randomly drawn test examples and RULER's string-match score.
From LOFT we take all five retrieval-augmented generation tasks: NQ, HotpotQA, MuSiQue, QAMPARI and QUEST.
Each task uses its first $25$ test questions, answered against one shared corpus, and LOFT's sub-span exact-match score.

\looseness=-1\textbf{Methods.}
Dense keeps the full KV cache and is the quality reference.
We compare \method{} with two query-agnostic top-$k$ eviction methods, KVzip~\citep{kvzip} and Fast KVzip~\citep{fastkvzip}, which keep the tokens with the highest reconstruction scores and learned gate scores, respectively.
To separate sampling from verification, we also evaluate \kvcomp{}($\mathcal{C}$=KVzip).
It draws the Poisson sample of Algorithm~\ref{alg:prefill} from KVzip scores and computes attention from the sample alone, with the unbiased estimator ($\lambda_i=1$) and without the certificate or \fetch{}.
\method{}($\mathcal{C}$, $\mathcal{A}$) adds the certificate and \fetch{} to a sample drawn from scorer $\mathcal{C}$ (KVzip or Fast KVzip), planning with sketch $\mathcal{A}$ (the 2-bit sketch, or a perfect sketch that uses the exact attention weights).
In the vLLM experiments, Dense (resident) keeps the full cache on the GPU, and Dense (offload) keeps it in host memory and transfers all of it at every decode step.

\begin{table*}[t]
  \centering
  \caption{Fetch statistics of \method{}($\mathcal{C}$=KVzip, $16\times$, RULER-32K, $\delta=0.1$). Decisions that fetch: share of decisions (one query head in one layer at one forward pass) that fetch at least one token. KV-head steps that fetch: the same per KV head, after merging the query heads that share it. Tokens per fetch: tokens fetched by a fetching decision, as \% of the context. KV read: fetched keys and values plus shortlist keys over all forward passes, as \% of the full cache.  At the default $\varepsilon_0=0.05$, three quarters of decisions fetch, yet each fetches only $4$--$6\%$ of the context and the overall read is $10.8\%$ of the cache. [Top] Detailed statistics for Qwen3-4B [Bottom] 2-bit sketch fetches across models.}
  \label{tab:fetch-stats}
  \small
  \input{image_assets/fetch_stats/fetch_stats}
\end{table*}

{\textbf{2-bit sketch implementation.}}
{The sketch stores a quarter of the key channels at $2$ bits each, costing $1.76\%$ of the full cache, paid from the GPU budget (Appendix~\ref{app:sketch}). At every forward pass it scores all non-resident tokens, and the $128$ highest-scoring keys per KV head are read from host memory and scored exactly. The certificate uses exact weights for these and the resident tokens, and sketch weights $\widehat w_i$ for the rest.}

\textbf{GPU memory budget.}
All methods hold the same amount of KV state on the GPU: $1/8$, $1/16$ or $1/32$ of the full cache, which we call $8\times$, $16\times$ and $32\times$ compression.
For \method{}, this budget covers everything it keeps on the GPU: $64$ sink tokens, the last $64$ context tokens, the sampled tokens and a $2$-bit key sketch (Appendix~\ref{app:sketch}).
The sketch costs $1.76\%$ of the full cache for all three models, so at $32\times$ the sink, local and sampled tokens share only $1.37\%$.
The full cache stays in host memory.
Unless stated otherwise, \method{} runs with $\varepsilon_0=0.05$, $\delta=0.1$ and a shortlist of $128$ tokens per KV head.

\textbf{Metrics.}
Quality is the mean task score of a benchmark on a $0$--$100$ scale, and we compare methods by their gap to Dense in points.
Host traffic is the KV data transferred from host memory: the keys and values of fetched tokens and the keys of the shortlist, as a share of the full cache.
We report it per forward pass, averaged over the question pass and all decode steps of an answer, and in Table~\ref{tab:punchline} per generated token.
An attention decision is the computation of one query head in one layer at one forward pass, and the fetch rate is the share of decisions that fetch at least one token.
In vLLM, we report decode throughput, time to first token (TTFT), time per output token (TPOT) and end-to-end latency per request. We now discuss the results in detail.

\subsection{Quality vs. Compression}
\label{sec:quality-compression}

Figure~\ref{fig:quality-compression} compares mean task scores across twelve model, benchmark, and context-length settings at matched GPU memory. We make the following observations. Detailed results are presented in Appendix~\ref{app:quality-full}.

\begin{itemize}[nosep, leftmargin=*]
    \looseness=-1\item Overall, \method{}($\mathcal{C}$=KVzip, $\mathcal{A}$=perfect sketch) {nearly matches the quality of dense execution ($99\%$ on average)}, even at extreme compression rates such as $32\times$ for GPU-resident cache.

    \looseness=-1\item With a 2-bit sketch, a naive representative of a sparse attention indexer, the gap widens slightly. However, it still significantly outperforms non-fetching static algorithms such as KVzip and Fast KVzip at high compression rates. We expect this gap to reduce further as the sketch improves.

    \looseness=-1\item \kvcomp{}, the pure compression component of \method{}, consistently outperforms existing KV compression methods and can be of independent interest to the community.

    \looseness=-1\item All non-fetching algorithms, including KVzip, Fast KVzip, and \kvcomp{}, suffer significant degradation at high compression rates, longer contexts, and on relatively difficult tasks such as LOFT. This is expected, since it is challenging, if not impossible, to expect a compressed KV cache to retain all the information required from the full context. This underlines our position that fetching is an essential component of a verified algorithm, without which verification is not possible.

    \looseness=-1\item The impact of compression is not uniform across tasks. For LOFT, which is a relatively difficult dataset for these models, static compression has a more severe impact than on RULER. However, the fetch algorithm is able to recover quality similarly on both datasets.

    \looseness=-1\item As context length increases, the quality of compression, even with \fetch{}, degrades. However, this is true only with an approximate sketch. When using a perfect sketch, we still recover the full dense quality. This implies that sketch quality degrades with context length and that better sketches are needed when dealing with longer contexts.
\end{itemize}

\begin{figure*}[!t]
    \centering
    \begin{minipage}[t]{0.48\textwidth}
        \centering
        \includegraphics[width=\linewidth]{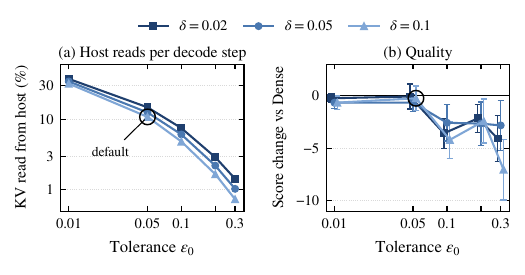}
        \caption{Effect of the tolerance $\varepsilon_0$ and failure probability $\delta$. \method{}($\mathcal{C}$=KVzip, $\mathcal{A}$=2-bit sketch) on Qwen3-4B, RULER-32K, $16\times$: KV read from host per decode step (a) and score change relative to Dense (b). $\varepsilon_0$ controls traffic, which falls about {$43\times$} from $\varepsilon_0=0.01$ to $0.3$, while $\delta$ changes it by less than $2\times$. Quality stays within one point of Dense up to $\varepsilon_0=0.05$, the default (circled), and degrades slightly beyond it.}
        \label{fig:eps-delta}
    \end{minipage}
    \hfill
    \begin{minipage}[t]{0.48\textwidth}
        \centering
        \includegraphics[width=\linewidth]{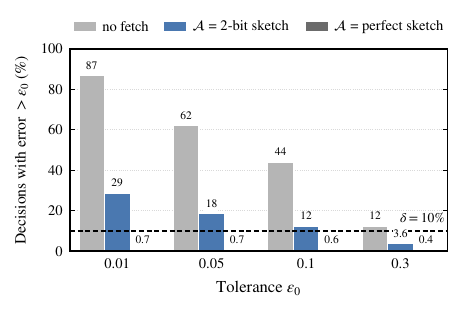}
        \caption{Error adherence of \method{}($\mathcal{C}$=KVzip) on Qwen3-4B, RULER-32K, $16\times$, $\delta=0.1$ (about one million attention decisions per setting). With a perfect sketch, under $1\%$ of decisions exceed $\varepsilon_0$, far below $\delta$ at every tolerance. The naive 2-bit sketch stays within $\delta$ only at $\varepsilon_0=0.3$, but remains far below the failure rate without fetching. Sketch error, not the certificate, is what limits adherence.}
        \label{fig:adherence}
    \end{minipage}
\end{figure*}

\subsection{Host Traffic with \method{}}
\label{sec:analysis}

Table~\ref{tab:fetch-stats} presents the fetching behavior of \fetch{} in detail for different models and values of $\varepsilon_0$, at $\delta=0.1$ . We make the following observations.

\begin{itemize}[nosep, leftmargin=*]
    \looseness=-1\item \fetch{} completely avoids accessing the CPU KV cache for an average of 20--25\% of queries (when using $\varepsilon_0=0.05$ ). This can be controlled using the tolerance parameters. For instance, by setting $\varepsilon_0=0.3$, we can avoid accessing the CPU for 65--67\% of queries. Thus, \method{} provides control over host traffic through its tolerance parameters.

    \looseness=-1\item While we still fetch for the majority of queries, the amount of data fetched as a fraction of the full KV cache is quite small. For instance, at $\varepsilon_0=0.05$, we read only $10.8\%$ of the KV cache on Qwen3-4B.

    \looseness=-1\item The amount of fetching varies across models. This is interesting because it points to differences in attention behavior across models. More importantly, we are able to capture these differences through our verification mechanism. For instance, we need to fetch approximately $3\times$ more on Ministral-3-8B than on Qwen3-4B.

    \looseness=-1\item {\textbf{Effect of grouped-query attention:} Query heads share KV heads (Qwen3-4B has $4$ query heads per KV head), and a KV head fetches the union of the tokens selected by its query heads. A KV head therefore fetches whenever any of its query heads does, which is why $75\%$ of decode decisions fetch but $93.5\%$ of KV-head steps end up fetching. Traffic is uneven across layers: the first three layers read $20$--$34\%$ of the cache per decode step, against a median of $7\%$.}

    \looseness=-1\item {\textbf{Effect of shrinkage:} Binary shrinkage also reduces fetching, because a lower-variance estimator needs fewer exact tokens to meet the same tolerance. On one RULER context at $16\times$ ($\varepsilon_0=0.05$, $\delta=0.1$), it lowers the mean fetch from about $6\%$ to $4\%$ of the context per decision and raises the share of decisions that need no fetch from $6\%$ to $21\%$.}

    \looseness=-1\item \textbf{Effect of $\delta$:} Lowering $\delta$ increases traffic by $1.2$--$1.9\times$ but has only a small consistent effect on quality at the current sample sizes.

\end{itemize}

\subsection{Better Decode Efficiency than Full KV Offloading}
\label{sec:systems-pareto}
\begin{figure}[!t]
  \centering
  \includegraphics[width=\columnwidth]{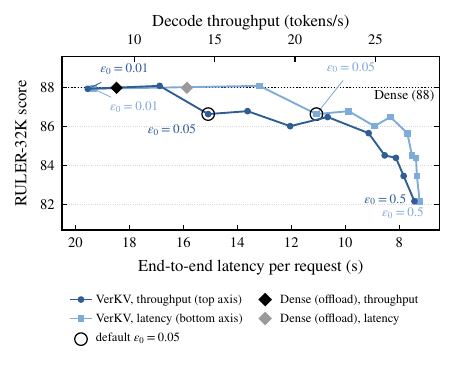}
  \caption{Quality vs.\ speed in vLLM. \method{}($\mathcal{C}$=Fast KVzip, $\mathcal{A}$=2-bit sketch) on Qwen3-4B, RULER-32K at $16\times$, sweeping $\varepsilon_0$ at $\delta=0.1$ against dense attention with the cache offloaded to host memory. At the default $\varepsilon_0=0.05$, \method{} decodes $1.6\times$ faster and cuts end-to-end latency by $5$~s at near-dense quality; larger $\varepsilon_0$ trades a few points of quality for up to $3\times$ throughput, with gains flattening once fetching becomes negligible.}
  \label{fig:sys-speed}
\end{figure}

Figure~\ref{fig:sys-speed} shows the latency-quality tradeoff of \method{}. A more detailed evaluation of the current version of our implementation is available in Appendix~\ref{app:vllm}. We make the following observations:
\begin{itemize}[leftmargin=*, nosep]
    \looseness=-1\item Throughput and latency: At the default $\varepsilon_0=0.05$, \method{} achieves a $1.6\times$ decoding speedup over Dense (offload), with only a small drop in quality. The tolerance parameter controls the quality-throughput tradeoff, with speedups reaching up to $3.1\times$ at $\varepsilon_0=0.5$.

    \looseness=-1\item Verification introduces kernel overhead even when fetching is negligible. At $\varepsilon_0=0.5$, decoding takes $36.6$~ms per token, compared to $5.8$~ms for Dense (resident), making token selection the primary remaining bottleneck. Optimizing our current implementation can lead to further improvements in efficiency. More details are in Appendix~\ref{app:vllm}.
\end{itemize}

\subsection{The Error Bound Guarantee}
\label{sec:adherence}
Figure~\ref{fig:adherence} shows how often the error bound holds. We make the following observations:
\begin{itemize}[leftmargin=*, nosep]
    \looseness=-1\item Guarantee adherence: With exact attention weights, the empirical failure rate is always under $0.7\%$, much below the target $\delta$ of $10\%$. Without fetching, $11\%$--$87\%$ of decisions violate the error tolerance, demonstrating the importance of verification and fetching.
    \looseness=-1\item Effect of sketch approximation: The $2$-bit sketch violates the target failure probability at stringent tolerances ($\varepsilon_0=0.01, 0.05, 0.1$), as sketch errors are not accounted for in the certificate. More accurate sketches or certificates that explicitly incorporate sketch error can close this gap.
\end{itemize}

%% file: image_assets/fetch_stats/fetch_stats.tex
\setlength{\tabcolsep}{3pt}
\begin{tabular*}{\textwidth}{@{\extracolsep{\fill}}ll|cc|cc|cc|c@{}}
\toprule
 & & \multicolumn{2}{c|}{Decisions that fetch (\%)} & \multicolumn{2}{c|}{KV-head steps that fetch (\%)} & \multicolumn{2}{c|}{Tokens per fetch (\% of context)} & KV read (\% of cache) \\
Sketch & $\varepsilon_0$ & Question & Decode & Question & Decode & Question & Decode & Overall \\
\midrule
2-bit & 0.01 & 94.1 & 92.2 & 100 & 99.5 & 17.4 & 13.9 & 31.9 \\
 & 0.05 & 79.7 & 75.3 & 100 & 93.5 & 5.9 & 4.1 & 10.8 \\
 & 0.1 & 68.6 & 62.7 & 100 & 84.1 & 2.8 & 1.7 & 4.8 \\
 & 0.3 & 34.6 & 32.9 & 97.0 & 56.7 & 0.5 & 0.3 & 0.7 \\
\midrule
perfect & 0.01 & 96.7 & 95.3 & 100 & 99.8 & 17.2 & 14.4 & 33.9 \\
 & 0.05 & 86.1 & 81.3 & 100 & 96.3 & 5.9 & 4.2 & 12.2 \\
 & 0.1 & 76.3 & 70.1 & 100 & 89.8 & 2.8 & 1.9 & 5.7 \\
 & 0.3 & 41.5 & 37.6 & 99.0 & 63.1 & 0.5 & 0.3 & 0.8 \\
\bottomrule
\end{tabular*}

\vspace{6pt}

\begin{tabular*}{\textwidth}{@{\extracolsep{\fill}}l|ccccc@{}}
\toprule
KV read per forward (\% of cache) & $\varepsilon_0=0.01$ & $\varepsilon_0=0.05$ & $\varepsilon_0=0.1$ & $\varepsilon_0=0.2$ & $\varepsilon_0=0.3$ \\
\midrule
Qwen3-4B & 31.8 & 10.8 & 4.8 & 1.7 & 0.7 \\
Qwen3.5-9B & 24.5 & 8.7 & 4.7 & 2.0 & 1.0 \\
Ministral-3-8B & 55.1 & 30.4 & 19.4 & 9.4 & 5.1 \\
\bottomrule
\end{tabular*}

%% file: sections/conclusion.tex
\section{Conclusion}
\label{sec:conclusion}

As context lengths grow, the KV cache can exceed GPU memory, and existing approaches, while promising, each carry specific limitations.
KV compaction regenerates the cache from a shorter context, which incurs substantial overhead and may alter the retained information.
KV offloading computes attention exactly, but its host-to-GPU transfers at every decoding step can dominate inference latency.
Query-agnostic KV compression keeps a small cache on the GPU, but provides no quality guarantee for an arbitrary query and degrades sharply at high compression rates.
Sparse attention identifies relevant entries using the current query, but still requires access to the full cache at every step.

We present \method{}, which computes verified attention in a hybrid GPU/CPU setting.
\method{} keeps a Poisson sample of the KV cache on the GPU and the full cache in host memory.
At every decode step, a certificate bounds the attention error of the sample, and tokens are fetched from host memory only while the bound exceeds the tolerance.
As a result, \method{} stays close to full-cache quality at compression rates where the quality of query-agnostic eviction collapses, while transferring a small fraction of the data that full offloading moves.

%% file: sections/appendix_theory.tex
\section{Implementation Details}
\label{app:impl}

\subsection{Key Sketch}
\label{app:sketch}
For each layer and KV head, the sketch keeps the $16$ RoPE channel pairs ($32$ of $128$ channels; a quarter of the channels for Qwen3.5-9B's $256$-dimensional heads) with the largest mean key energy. RoPE rotates within each pair, so a pair's share of $\langle q,k\rangle$ does not depend on position and the choice needs no queries.
Each kept channel is {quantized} to $2$ bits, uniformly between the minimum and maximum of each group of $128$ tokens, with an fp16 zero point and scale per group. The sketch costs $1.76\%$ of the full cache and is paid from the GPU budget.
At every forward pass, the sketch scores all non-resident tokens. The $128$ tokens per KV head with the largest sketch score over that head's query rows form the shortlist: their keys are read from host memory and scored exactly. The certificate uses exact weights for resident and shortlisted tokens and sketch weights $\widehat w_i$ for the rest.

\subsection{Binary Shrinkage}
\label{app:binary-shrinkage}
The implemented estimator uses a simpler version of the unique minimizer of the MSE~\eqref{eq:mse}: instead of solving for the clipped coefficients $\lambda_i\in[r_i,1]$ of Definition~\ref{def:estimator}, each coefficient takes one of the two endpoints, $\lambda_i\in\{1,r_i\}$.
With $\lambda_i=1$, a sampled token keeps its unbiased weight $w_i/r_i$; with $\lambda_i=r_i$, it receives its ordinary weight $w_i$, which reduces its variance at the cost of a known bias $a_i=\widehat w_i(1-r_i)$.
Moving token $i$ from $\lambda_i=1$ to $\lambda_i=r_i$ changes the MSE~\eqref{eq:mse} by exactly $a_i\big(2\beta-\widehat w_i(1-r_i)/r_i\big)$, where $\beta$ is the bias already accumulated.
We therefore visit the tokens of $L$ in decreasing $\widehat w_i(1-r_i)/r_i$ and set $\lambda_i=r_i$ while $\widehat w_i(1-r_i)/r_i>2\beta$.
Along this order the first quantity decreases and $\beta$ increases, so the point where we stop is the MSE minimum over all prefixes of the order.

Figure~\ref{fig:binary-shrinkage-main} traces the MSE along this order for one query row (RULER cwe, layer $21$, decode step $21$, $|L|=44{,}549$), from all $\lambda_i=1$ (the unbiased estimator) to all $\lambda_i=r_i$, which is $1.62\times$ worse.
With exact weights, the minimum is at $52$ tokens ($0.12\%$ of $L$): $0.54\times$ the MSE of the unbiased estimator.
The deployed algorithm computes the order and $\beta$ from the $2$-bit sketch weights $\widehat w_i$. Because the sketch is imperfect, its estimate of the minimum falls short ($16$ tokens) and partly selects different tokens, but it still reduces the MSE to $0.69\times$.

\subsection{vLLM Implementation}
\label{app:vllm}
\subsubsection{Token-level offloading and block compaction}
We implement token-level offloading inside vLLM. After prefill, we score
the context with Fast KVzip and sample, per layer and KV head. The sampled context
entries stay resident on the GPU; the rest are offloaded to host
memory. Let $L$, $H$, $P$ and $C$ denote the number of layers, KV heads,
tokens per page and the head dimension. vLLM's paged cache allocates KV
memory in blocks, so offloading individual entries only leaves holes;
to reclaim device memory, the surviving entries must be compacted into
fewer blocks.

vLLM stores all layers' pages in a single backing buffer, which we view as
a flat array of \emph{cells}: a cell is the $2C$-element K/V row of one
token in one head of one layer, with a single global index. After the
sampling round, the $T$ surviving entries of a request (across all layers
and heads) are packed contiguously, in (layer, head, position) order, into
the first $n_b = \lceil T/(LHP) \rceil$ blocks of that request, and the
remaining blocks are returned to vLLM's block pool. Each request records an
offset $\Delta = L_g - n_b P$ ($L_g$: logical length); the scheduler and
model runner subtract $\Delta$ from sequence lengths and slot mappings, so
newly generated tokens are written in vLLM's native layout directly after
the packed region, while RoPE still uses logical positions.

Offloaded entries are copied once, in chunks, into a pinned host pool of shape $[N, 2C]$ with
$N = L\,H\,N_{\mathrm{cpu}}$; free rows are managed by a stack. Two maps
per request locate every entry: $\mathrm{cell}[l,p,h]$ gives the global
cell of a resident entry ($-1$ if offloaded), and
$\mathrm{cpu\_row}[l,p,h]$ gives its host row. A per-(layer, head) list of
resident positions, extended as tokens are generated, defines the entries
that are scored and attended to at every step. Currently we enforce synchronous scheduling.

\subsubsection{Recall implementation}

\paragraph{Selection.} The decide kernel assigns a thread-block cluster
(up to 4 blocks, communicating through shared memory) to each
(request, query head). A radix
\emph{select} (8-bit digit histograms, merged across the cluster) finds
the top $K{=}4096$ candidates, and only these are moved to shared memory and sorted, each block sorting a quarter
with a block radix sort followed by a rank-based merge. The fetch count
$k^\star$ is computed in the same kernel: the bound is evaluated for all $k \le K$ at once with suffix
scans over the sorted head. The $\lambda$ grid split across the cluster,
and lower-ranked entries aggregated without sorting. Rows that find no valid $k$  are flagged on the
device and recomputed by a kernel that sorts the full
context in global memory and overwrites the results in place.

\paragraph{Fetching.} Fetching is per layer and only covers the
offloaded entries of the selected set; all requests and heads of a
layer are served by one batched launch. GPU threads copy the required host rows directly over UVA (zero-copy; no \texttt{cudaMemcpy} or CPU involvement) into a staging area on host. A device-side map keyed by (request, KV head, position) de-duplicates copies across the query heads sharing a KV head and reuses rows staged in earlier steps.

\paragraph{Attention.} Attention is a single sparse paged-decode kernel
over global cell indices. Importance weights
enter as an additive $\log w$ on the logits ($w = 1/r$ for sampled
entries, $1$ otherwise). Each row uses a fixed-width list (resident entries
plus up to $K$ fetched ones); fallback rows that need more use a second
list, and the two partial softmaxes are merged exactly by log-sum-exp.

\subsubsection{Metrics}
\begin{figure*}[!t]
  \centering
  \includegraphics[width=\textwidth]{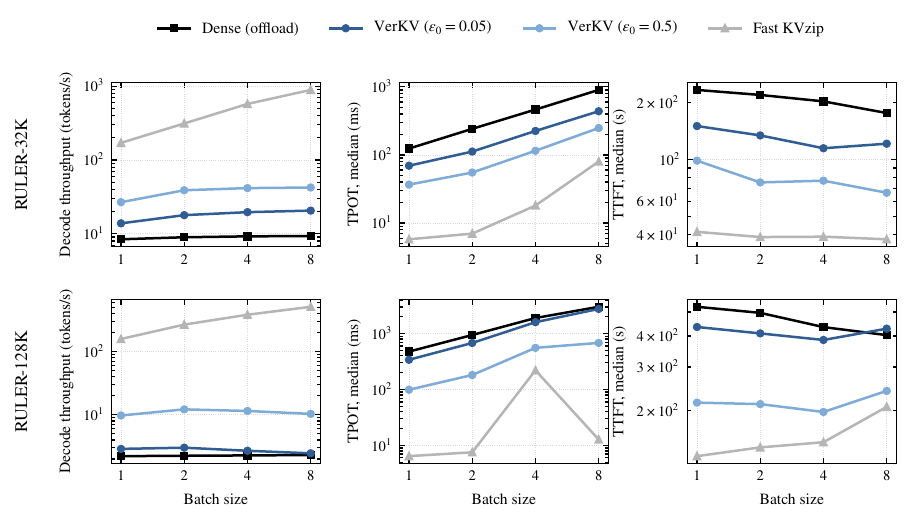}
  \caption{Decode throughput, time per output token and time to first token vs.\ batch size in vLLM (Qwen3-4B, RULER-32K and RULER-128K, $16\times$, $\delta=0.05$). At 32K, \method{} stays ahead of Dense (offload) at every batch size, and its throughput grows from $13.8$ to $20.6$ tokens/s at $\varepsilon_0=0.05$ and from $26.8$ to $42.4$ tokens/s at $\varepsilon_0=0.5$, while Dense (offload) is bound by PCIe and stays near $9$ tokens/s. The gain from batching is small because the decide kernel saturates shared memory and runs the requests of a batch one after another. At 128K, \method{} at $\varepsilon_0=0.05$ approaches Dense (offload) as the batch grows, whereas $\varepsilon_0=0.5$ keeps a $4$--$5\times$ advantage. Fast KVzip is fastest throughout but loses most of its quality (Table~\ref{tab:punchline}).}
  \label{fig:batch-size}
\end{figure*}

\begin{figure}[!htb]
  \centering
  \includegraphics[width=\columnwidth]{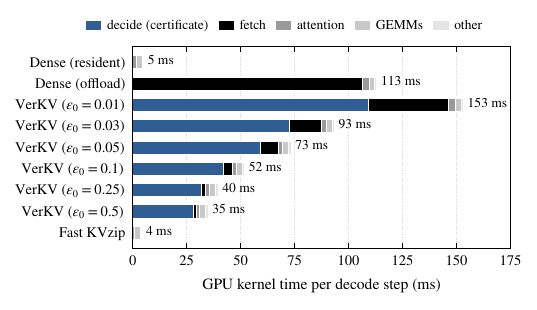}
  \caption{GPU kernel time per decode step, split by phase (Qwen3-4B, RULER-32K, $16\times$, batch size $1$, $\delta=0.05$). Dense (offload) spends almost all of its step fetching the cache over PCIe. \method{} spends most of its step in the decide kernel that computes the certificate and the fetch set, and only $1$--$37$~ms fetching; at the default $\varepsilon_0=0.05$, fetching takes $8$~ms against $107$~ms for Dense (offload). Faster decide kernels are therefore the main lever for further speedups.}
  \label{fig:decode_profile}
\end{figure}

\begin{figure}[!htb]
  \centering
  \includegraphics[width=\columnwidth]{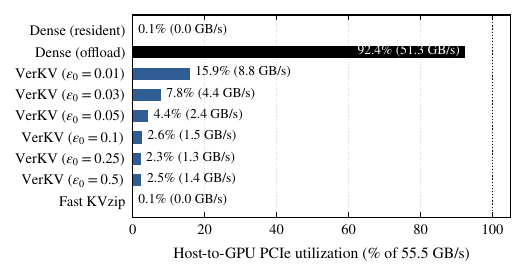}
  \caption{Host-to-GPU PCIe utilization during decode (Qwen3-4B, RULER-32K, $16\times$, batch size $1$, $\delta=0.05$; measured link bandwidth $55.5$~GB/s). Dense (offload) saturates the link at $92\%$, while \method{} uses $2$--$16\%$ of it ($4.4\%$ at the default $\varepsilon_0=0.05$), leaving bandwidth for other requests or larger contexts.}
  \label{fig:pcie}
\end{figure}

Figure~\ref{fig:decode_profile} shows that Dense (offload) spends most of its time in the fetch phase, whereas \method{} is bound by the algorithm's kernel to decide which entries to fetch. The fetch time of \method{} is much smaller than that of Dense (offload).\\
We can also see in Figure~\ref{fig:pcie} that Dense (offload) puts a significantly higher pressure on the PCIe link, saturating it, whereas \method{} needs much less PCIe bandwidth.\\
We study the effect of batch size on throughput, TPOT and TTFT (Figure~\ref{fig:batch-size}). We observe that the throughput of \method{} does not increase much as batch size increases since one batch of \method{} saturates the shared memory (as it involves sorting and topk computations) effectively making the decide kernel execution serial for batches.

\section{Proofs}
\label{app:proofs}

\begin{proof}[Proof of Theorem~1]
Fix the query, the sets $H,L$, the weights $w_i$,
and the inclusion probabilities $r_i$.
The weights satisfy $w_i>0$ for every $i\in L$.
Moreover, $0<r_i<1$ on $L$, and the variables
$Z_i\sim\operatorname{Bernoulli}(r_i)$ are independent.
All coefficients considered below are chosen independently
of the sampling outcomes.

If $L=\varnothing$, the estimator is exact and there
are no coefficients to optimize. Henceforth assume
$L\ne\varnothing$.

For any coefficients satisfying $r_i\le\lambda_i\le1$,
consider the estimator
\[
\widehat D(\lambda)
=
\sum_{i\in H}w_i
+
\sum_{i\in L}\frac{Z_i\lambda_iw_i}{r_i}.
\]
Its bias is
\[
\mathbb E[\widehat D(\lambda)]-D
=
-\sum_{i\in L}w_i(1-\lambda_i).
\]
By independence of the sampling indicators,
\[
\operatorname{Var}(\widehat D(\lambda))
=
\sum_{i\in L}
\left(\frac{\lambda_iw_i}{r_i}\right)^2
r_i(1-r_i)
=
\sum_{i\in L}\lambda_i^2w_i^2\frac{1-r_i}{r_i}.
\]
Therefore,
\[
\begin{aligned}
\mathbb E\!\left[(\widehat D(\lambda)-D)^2\right]
&=
\left(\sum_{i\in L} w_i(1-\lambda_i)\right)^2\\
&\quad+
\sum_{i\in L}
\lambda_i^2 w_i^2\frac{1-r_i}{r_i}.
\end{aligned}
\]
This is exactly the objective in~\eqref{eq:mse}.
We denote this objective by $f(\lambda)$.

The feasible set, defined by $r_i\le\lambda_i\le1$
for all $i\in L$, is nonempty, compact, and convex.
Since $f$ is continuous, a minimizer exists.
Furthermore, for every nonzero vector
$u=(u_i)_{i\in L}$,
\[
\begin{aligned}
u^\top\nabla^2 f(\lambda)u
&=
2\left(\sum_{i\in L} w_i u_i\right)^2\\
&\quad+
2\sum_{i\in L}
 w_i^2\frac{1-r_i}{r_i}u_i^2
>0.
\end{aligned}
\]
The strict inequality follows from
$ w_i>0$ and $0<r_i<1$.
Thus $f$ is strictly convex, and its minimizer
$\lambda^\star$ is unique.

It remains to identify this minimizer.
Let
\[
\beta^\star
=
\sum_{i\in L} w_i(1-\lambda_i^\star).
\]
Differentiating the objective gives
\[
\begin{aligned}
\frac{\partial f}{\partial\lambda_i}(\lambda^\star)
&=
-2 w_i\beta^\star
+
2 w_i^2\frac{1-r_i}{r_i}\lambda_i^\star\\
&=
2 w_i^2\frac{1-r_i}{r_i}
\left(
\lambda_i^\star
-
\frac{\beta^\star r_i}
{ w_i(1-r_i)}
\right).
\end{aligned}
\]

The first-order optimality conditions for the
interval constraints are
\[
\frac{\partial f}{\partial\lambda_i}(\lambda^\star)
\begin{cases}
\ge 0, & \lambda_i^\star=r_i,\\
=0, & r_i<\lambda_i^\star<1,\\
\le 0, & \lambda_i^\star=1.
\end{cases}
\]
Necessity follows by considering feasible changes
to each coordinate.
These conditions are also sufficient: for any
feasible $\lambda$, convexity gives
\[
f(\lambda)
\ge
f(\lambda^\star)
+
\sum_{i\in L}
\frac{\partial f}{\partial\lambda_i}(\lambda^\star)
(\lambda_i-\lambda_i^\star)
\ge
f(\lambda^\star).
\]
The final inequality follows from the sign conditions
above and the interval constraints.
Since
$2 w_i^2(1-r_i)/r_i>0$,
these optimality conditions are equivalent to
\[
\lambda_i^\star
=
\operatorname{clip}_{[r_i,1]}
\left(
\frac{\beta^\star r_i}
{ w_i(1-r_i)}
\right),
\qquad i\in L.
\]
Together with
\[
\beta^\star
=
\sum_{i\in L} w_i(1-\lambda_i^\star),
\]
these are precisely the coupled equations in
Definition~\ref{def:estimator}.

Existence of the minimizer therefore establishes
existence of a solution to those equations.
Conversely, every solution satisfies the sufficient
optimality conditions above and hence minimizes $f$.
Strict convexity makes the coefficient vector unique,
and its defining sum then makes $\beta^\star$ unique.
Thus Definition~\ref{def:estimator} specifies the unique
MSE-minimizing coefficient vector over the stated
constraints.
\end{proof}

\begin{proof}[Proof of Lemma~\ref{lemma:bernstien}]
Fix the query, the sets $H,L$, the weights $w_i$,
and the inclusion probabilities $r_i$.
The coefficients $\lambda_i$ in
Definition~\ref{def:estimator} are determined without
using the sampling outcomes and satisfy
$r_i\le\lambda_i\le1$.
All probabilities and expectations below are over
the independent variables
$Z_i\sim\operatorname{Bernoulli}(r_i)$, $i\in L$.

Under exact sketch recovery,
\[ D
=
\sum_{i\in H}w_i+\sum_{i\in L} w_i
>0.
\]
If $L=\varnothing$, then $\widehat D=D$ and the
claim is immediate, taking empty sums and maxima
to be zero. Henceforth assume $L\ne\varnothing$.

For each $i\in L$, define the centered contribution
\[
X_i
=
\frac{\lambda_iw_i}{r_i}(Z_i-r_i).
\]
These variables are independent and satisfy
$\mathbb E[X_i]=0$. Moreover,
\[
\begin{aligned}
\widehat D-D
&=
\sum_{i\in L}
\left(\frac{Z_i\lambda_iw_i}{r_i}-w_i\right)\\
&=
-\sum_{i\in L}w_i(1-\lambda_i)
+\sum_{i\in L}X_i\\
&=
-\beta(L)+\sum_{i\in L}X_i.
\end{aligned}
\]
Since $\lambda_i\le1$, we have
$\beta(L)\ge0$.

The variance of each centered contribution is
\[
\mathbb E[X_i^2]
=
\left(\frac{\lambda_iw_i}{r_i}\right)^2
r_i(1-r_i)
=
\lambda_i^2w_i^2\frac{1-r_i}{r_i}.
\]
Consequently,
\[
\sum_{i\in L}\mathbb E[X_i^2]
=
 V(L).
\]

When $Z_i=0$, we have $X_i=-\lambda_iw_i$;
when $Z_i=1$, we have
$X_i=\lambda_iw_i(1-r_i)/r_i$.
Therefore,
\[
|X_i|
\le
\frac{\lambda_iw_i}{r_i}
\max\{r_i,1-r_i\}
\le
 b(L).
\]

Applying the two-sided Bernstein inequality gives,
for every $t>0$,
\[
\Pr\!\left(
\left|\sum_{i\in L}X_i\right|>t
\right)
\le
2\exp\!\left(
-\frac{t^2}
{2\bigl( V(L)+ b(L)t/3\bigr)}
\right).
\]
Let $l_\delta=\log(2/\delta)$ and set
\[
t_\delta
=
\frac{ b(L)l_\delta}{3}
+
\sqrt{
\left(\frac{ b(L)l_\delta}{3}\right)^2
+2 V(L)l_\delta
}.
\]
This is the positive solution of
\[
t_\delta^2
=
2l_\delta
\left(
 V(L)+\frac{ b(L)t_\delta}{3}
\right).
\]
Substitution into Bernstein's inequality yields
\[
\Pr\!\left(
\left|\sum_{i\in L}X_i\right|>t_\delta
\right)
\le 2e^{-l_\delta}
=\delta.
\]

Finally, the error decomposition and
$\beta(L)\ge0$ imply
\[
|\widehat D-D|
\le
\beta(L)
+
\left|\sum_{i\in L}X_i\right|.
\]
Thus, with probability at least $1-\delta$,
\[
\frac{|\widehat D-D|}{D}
\le
\frac{\beta(L)+t_\delta}{D}
=
\frac{\beta(L)+t_\delta}{ D}
=
\widehat\varepsilon(L),
\]
which proves the claim.
\end{proof}

\begin{proof}[Proof of Theorem 2]
In general, this is in line with the proof of Lemma~~\ref{lemma:bernstien}. First, we fix the query, the sets $H,L$, the weights $w_i$,
and the inclusion probabilities $r_i$.
The coefficients $\lambda_i\in[r_i,1]$ and the
ordering $L_g$ are chosen without using the sampling
results and also remain fixed.
All probabilities below are over the independent
variables $Z_i\sim\operatorname{Bernoulli}(r_i)$,
$i\in L$.
If the coefficients or ordering use additional
randomness independent of this sample, the argument
below is conditional on that randomness; averaging
then gives the same conclusion.

We interpret $L_g[k:]$ as the set remaining after
removing the first $k$ indices from the ordering.
In particular, $L_g[0:]=L$ and
$L_g[|L|:]=\varnothing$.
Throughout the proof, empty sums and maxima are zero,
and every certificate uses the denominator
\[
D=\sum_{i\in H}w_i+\sum_{i\in L}w_i>0.
\]

First, the stopping index exists.
Since $\lambda_i\le1$, every summand in $\beta(L)$
is nonnegative. The summands in $V(L)$ and the
quantities defining $b(L)$ are also nonnegative.
Consequently, with the coefficients held fixed,
$\beta(L_g[k:])$, $V(L_g[k:])$, and $b(L_g[k:])$
are all nonincreasing in $k$.
The certificate in~\eqref{eq:epsiloncomputation} is nondecreasing
in each of these three quantities, so
$\widehat\varepsilon(L_g[k:])$ is nonincreasing in $k$.
Moreover,
\[
\widehat\varepsilon(L_g[|L|:])
=\widehat\varepsilon(\varnothing)=0<\epsilon.
\]
Thus
\[
k^\star
=
\min\left\{
k\in\{0,\ldots,|L|\}:
\widehat\varepsilon(L_g[k:])<\epsilon
\right\}
\]
is well defined.

Each certificate is computed from the fixed weights,
inclusion probabilities, coefficients, and ordering.
Therefore $k^\star$ and the selected set
$L'=L_g[k^\star:]$ do not depend on the sampling
outcomes.
In particular, selecting $L'$ does not change the
independent Bernoulli distributions of the variables
$Z_i$, $i\in L'$.

The algorithm makes every contribution in
$L\setminus L'$ exact: tokens already in $S$ are
reused, and the remaining tokens are fetched.
Hence the returned estimator is
\[
\widehat D
=
\sum_{i\in H}w_i
+\sum_{i\in L\setminus L'}w_i
+\sum_{i\in L'}\frac{Z_i\lambda_iw_i}{r_i}.
\]
If $L'=\varnothing$, then $\widehat D=D$ and the
claimed probability bound follows immediately.
Assume henceforth that $L'\ne\varnothing$.

For each $i\in L'$, define
\[
X_i=\frac{\lambda_iw_i}{r_i}(Z_i-r_i).
\]
These variables are independent and centered, and
\[
\mathbb E[X_i^2]
=
\lambda_i^2w_i^2\frac{1-r_i}{r_i}.
\]
Their two possible values are $-\lambda_iw_i$ and
$\lambda_iw_i(1-r_i)/r_i$, so
\[
|X_i|
\le
\frac{\lambda_iw_i}{r_i}\max\{r_i,1-r_i\}
\le b(L').
\]
Thus
\[
\sum_{i\in L'}\mathbb E[X_i^2]=V(L').
\]
Subtracting the exact denominator gives
\[
\begin{aligned}
\widehat D-D
&=
\sum_{i\in L'}
\left(\frac{Z_i\lambda_iw_i}{r_i}-w_i\right)\\
&=
-\sum_{i\in L'}w_i(1-\lambda_i)
+\sum_{i\in L'}X_i\\
&=
-\beta(L')+\sum_{i\in L'}X_i.
\end{aligned}
\]

Let $l_\delta=\log(2/\delta)$ and define
\[
t_\delta
=
\frac{b(L')l_\delta}{3}
+
\sqrt{
\left(\frac{b(L')l_\delta}{3}\right)^2
+2V(L')l_\delta
}.
\]
Since $L'\ne\varnothing$, $w_i>0$, and
$0<r_i\le\lambda_i\le1$ with $r_i<1$,
we have $V(L')>0$ and hence $t_\delta>0$.
By construction,
\[
t_\delta^2
=
2l_\delta
\left(V(L')+\frac{b(L')t_\delta}{3}\right).
\]
The two-sided Bernstein inequality therefore yields
\[
\begin{aligned}
\Pr\!\left(
\left|\sum_{i\in L'}X_i\right|>t_\delta
\right)
&\le
2\exp\!\left(
-\frac{t_\delta^2}
{2\left(V(L')+b(L')t_\delta/3\right)}
\right)\\
&=2e^{-l_\delta}
=\delta.
\end{aligned}
\]

Since $\beta(L')\ge0$, with probability at least
$1-\delta$ we consequently have
\[
\begin{aligned}
\frac{|\widehat D-D|}{D}
&\le
\frac{\beta(L')+
      \left|\sum_{i\in L'}X_i\right|}{D}\\
&\le
\frac{\beta(L')+t_\delta}{D}
=
\widehat\varepsilon(L')
<
\epsilon.
\end{aligned}
\]
It follows that
\[
\Pr\!\left(
\frac{|\widehat D-D|}{D}>\epsilon
\right)\le\delta.
\]
No union bound over the candidate sets is needed:
the selected set $L'$ is determined independently
of the Bernoulli sample.
\end{proof}

\section{Full Per-Task Quality Tables}
\label{app:quality-full}
The tables below give the per-task scores ($0$--$100$) behind Figure~\ref{fig:quality-compression} for every model, benchmark, context length and compression rate, one table per model, benchmark and context. KVz / F-KVz: KVzip / Fast KVzip eviction; PC+K: \kvcomp{}($\mathcal{C}$=KVzip) ($\lambda_i=1$, no fetch); VKV+K / VKV+F: \method{}($\mathcal{C}$=KVzip / Fast KVzip, $\mathcal{A}$=2-bit sketch). RULER tasks: multi-key (MK3) and multi-value (MV) needle retrieval, common-word extraction (CWE), question answering (QA1); task sets as in Section~\ref{sec:setup}.

\input{image_assets/quality/quality_per_task_appendix}

%% file: image_assets/quality/quality_per_task_appendix.tex
\noindent\resizebox{\columnwidth}{!}{%
\begin{tabular}{llrrrrrr}
\toprule
\multicolumn{8}{c}{\textbf{Qwen3-4B, RULER-32K}} \\
\midrule
Rate & Task & Dense & KVz & F-KVz & PC+K & VKV+K & VKV+F \\
\midrule
8$\times$ & MK3 & 100.0 & 68.0 & 90.0 & 98.0 & 100.0 & 100.0 \\
 & MV & 100.0 & 81.0 & 68.0 & 95.0 & 100.0 & 100.0 \\
 & CWE & 75.6 & 63.0 & 52.6 & 67.8 & 72.6 & 75.6 \\
 & QA1 & 76.0 & 64.0 & 60.0 & 72.0 & 72.0 & 74.0 \\
 & \textit{Mean} & 87.9 & 69.0 & 67.7 & 83.2 & 86.2 & 87.4 \\
\midrule
16$\times$ & MK3 & 100.0 & 4.0 & 28.0 & 90.0 & 100.0 & 100.0 \\
 & MV & 100.0 & 36.0 & 17.0 & 53.0 & 100.0 & 100.0 \\
 & CWE & 75.6 & 2.2 & 7.6 & 50.0 & 71.4 & 70.6 \\
 & QA1 & 76.0 & 34.0 & 28.0 & 44.0 & 74.0 & 74.0 \\
 & \textit{Mean} & 87.9 & 19.1 & 20.1 & 59.2 & 86.3 & 86.2 \\
\midrule
32$\times$ & MK3 & 100.0 & 0.0 & 0.0 & 0.0 & 100.0 & 100.0 \\
 & MV & 100.0 & 2.5 & 2.0 & 1.0 & 99.5 & 100.0 \\
 & CWE & 75.6 & 0.4 & 0.4 & 29.4 & 69.8 & 65.0 \\
 & QA1 & 76.0 & 18.0 & 10.0 & 28.0 & 76.0 & 76.0 \\
 & \textit{Mean} & 87.9 & 5.2 & 3.1 & 14.6 & 86.3 & 85.2 \\
\bottomrule
\end{tabular}}
\par\vspace{1.2em}

\noindent\resizebox{\columnwidth}{!}{%
\begin{tabular}{llrrrrrr}
\toprule
\multicolumn{8}{c}{\textbf{Qwen3-4B, RULER-128K}} \\
\midrule
Rate & Task & Dense & KVz & F-KVz & PC+K & VKV+K & VKV+F \\
\midrule
8$\times$ & MK3 & 88.0 & 2.0 & 14.0 & 78.0 & 88.0 & 88.0 \\
 & MV & 100.0 & 76.5 & 71.0 & 99.0 & 100.0 & 100.0 \\
 & CWE & 4.2 & 8.2 & 2.0 & 3.4 & 5.2 & 4.8 \\
 & QA1 & 68.0 & 48.0 & 46.0 & 72.0 & 70.0 & 66.0 \\
 & \textit{Mean} & 65.0 & 33.7 & 33.2 & 63.1 & 65.8 & 64.7 \\
\midrule
16$\times$ & MK3 & 88.0 & 0.0 & 0.0 & 56.0 & 80.0 & 72.0 \\
 & MV & 100.0 & 34.0 & 22.0 & 78.0 & 100.0 & 100.0 \\
 & CWE & 4.2 & 2.4 & 0.2 & 3.6 & 3.8 & 4.2 \\
 & QA1 & 68.0 & 26.0 & 24.0 & 52.0 & 68.0 & 70.0 \\
 & \textit{Mean} & 65.0 & 15.6 & 11.6 & 47.4 & 63.0 & 61.5 \\
\midrule
32$\times$ & MK3 & 88.0 & 0.0 & 0.0 & 0.0 & 60.0 & 66.0 \\
 & MV & 100.0 & 1.5 & 2.5 & 5.0 & 100.0 & 99.5 \\
 & CWE & 4.2 & 0.8 & 0.6 & 5.2 & 5.4 & 3.6 \\
 & QA1 & 68.0 & 18.0 & 14.0 & 26.0 & 66.0 & 68.0 \\
 & \textit{Mean} & 65.0 & 5.1 & 4.3 & 9.1 & 57.9 & 59.3 \\
\bottomrule
\end{tabular}}
\par\vspace{1.2em}

\noindent\resizebox{\columnwidth}{!}{%
\begin{tabular}{llrrrrrr}
\toprule
\multicolumn{8}{c}{\textbf{Qwen3-4B, LOFT-32K}} \\
\midrule
Rate & Task & Dense & KVz & F-KVz & PC+K & VKV+K & VKV+F \\
\midrule
8$\times$ & NQ & 40.0 & 12.0 & 32.0 & 48.0 & 28.0 & 24.0 \\
 & HotpotQA & 44.0 & 52.0 & 44.0 & 60.0 & 40.0 & 44.0 \\
 & MuSiQue & 28.0 & 20.0 & 28.0 & 8.0 & 24.0 & 28.0 \\
 & QAMPARI & 48.0 & 0.0 & 0.0 & 24.0 & 40.0 & 48.0 \\
 & QUEST & 16.0 & 0.0 & 8.0 & 8.0 & 16.0 & 16.0 \\
 & \textit{Mean} & 35.2 & 16.8 & 22.4 & 29.6 & 29.6 & 32.0 \\
\midrule
16$\times$ & NQ & 40.0 & 16.0 & 8.0 & 44.0 & 32.0 & 32.0 \\
 & HotpotQA & 44.0 & 20.0 & 24.0 & 44.0 & 48.0 & 48.0 \\
 & MuSiQue & 28.0 & 0.0 & 4.0 & 0.0 & 28.0 & 36.0 \\
 & QAMPARI & 48.0 & 0.0 & 0.0 & 0.0 & 48.0 & 44.0 \\
 & QUEST & 16.0 & 0.0 & 0.0 & 0.0 & 16.0 & 12.0 \\
 & \textit{Mean} & 35.2 & 7.2 & 7.2 & 17.6 & 34.4 & 34.4 \\
\midrule
32$\times$ & NQ & 40.0 & 0.0 & 0.0 & 20.0 & 32.0 & 32.0 \\
 & HotpotQA & 44.0 & 4.0 & 20.0 & 0.0 & 48.0 & 52.0 \\
 & MuSiQue & 28.0 & 0.0 & 0.0 & 0.0 & 24.0 & 24.0 \\
 & QAMPARI & 48.0 & 0.0 & 0.0 & 0.0 & 48.0 & 44.0 \\
 & QUEST & 16.0 & 0.0 & 0.0 & 0.0 & 20.0 & 16.0 \\
 & \textit{Mean} & 35.2 & 0.8 & 4.0 & 4.0 & 34.4 & 33.6 \\
\bottomrule
\end{tabular}}
\par\vspace{1.2em}

\noindent\resizebox{\columnwidth}{!}{%
\begin{tabular}{llrrrrrr}
\toprule
\multicolumn{8}{c}{\textbf{Qwen3-4B, LOFT-128K}} \\
\midrule
Rate & Task & Dense & KVz & F-KVz & PC+K & VKV+K & VKV+F \\
\midrule
8$\times$ & NQ & 20.0 & 24.0 & 4.0 & 12.0 & 20.0 & 20.0 \\
 & HotpotQA & 32.0 & 36.0 & 44.0 & 32.0 & 40.0 & 32.0 \\
 & MuSiQue & 8.0 & 4.0 & 0.0 & 4.0 & 0.0 & 12.0 \\
 & QAMPARI & 8.0 & 0.0 & 0.0 & 8.0 & 8.0 & 12.0 \\
 & QUEST & 8.0 & 4.0 & 4.0 & 4.0 & 4.0 & 8.0 \\
 & \textit{Mean} & 15.2 & 13.6 & 10.4 & 12.0 & 14.4 & 16.8 \\
\midrule
16$\times$ & NQ & 20.0 & 4.0 & 8.0 & 4.0 & 16.0 & 20.0 \\
 & HotpotQA & 32.0 & 4.0 & 24.0 & 40.0 & 28.0 & 36.0 \\
 & MuSiQue & 8.0 & 0.0 & 8.0 & 8.0 & 4.0 & 4.0 \\
 & QAMPARI & 8.0 & 0.0 & 0.0 & 0.0 & 8.0 & 4.0 \\
 & QUEST & 8.0 & 0.0 & 0.0 & 4.0 & 4.0 & 4.0 \\
 & \textit{Mean} & 15.2 & 1.6 & 8.0 & 11.2 & 12.0 & 13.6 \\
\midrule
32$\times$ & NQ & 20.0 & 0.0 & 0.0 & 0.0 & 20.0 & 16.0 \\
 & HotpotQA & 32.0 & 12.0 & 8.0 & 28.0 & 36.0 & 40.0 \\
 & MuSiQue & 8.0 & 0.0 & 0.0 & 4.0 & 4.0 & 4.0 \\
 & QAMPARI & 8.0 & 0.0 & 0.0 & 0.0 & 4.0 & 8.0 \\
 & QUEST & 8.0 & 0.0 & 0.0 & 0.0 & 12.0 & 4.0 \\
 & \textit{Mean} & 15.2 & 2.4 & 1.6 & 6.4 & 15.2 & 14.4 \\
\bottomrule
\end{tabular}}
\par\vspace{1.2em}

\noindent\resizebox{\columnwidth}{!}{%
\begin{tabular}{llrrrrrr}
\toprule
\multicolumn{8}{c}{\textbf{Qwen3.5-9B, RULER-32K}} \\
\midrule
Rate & Task & Dense & KVz & F-KVz & PC+K & VKV+K & VKV+F \\
\midrule
8$\times$ & MK3 & 100.0 & 0.0 & 14.0 & 0.0 & 100.0 & 100.0 \\
 & MV & 100.0 & 41.0 & 2.5 & 4.0 & 100.0 & 100.0 \\
 & CWE & 98.4 & 30.6 & 55.4 & 82.4 & 90.8 & 92.6 \\
 & QA1 & 90.0 & 36.0 & 16.0 & 48.0 & 90.0 & 92.0 \\
 & \textit{Mean} & 97.1 & 26.9 & 22.0 & 33.6 & 95.2 & 96.2 \\
\midrule
16$\times$ & MK3 & 100.0 & 0.0 & 0.0 & 0.0 & 100.0 & 96.0 \\
 & MV & 100.0 & 1.0 & 2.0 & 0.0 & 100.0 & 100.0 \\
 & CWE & 98.4 & 1.4 & 40.4 & 74.0 & 90.6 & 92.4 \\
 & QA1 & 90.0 & 12.0 & 16.0 & 28.0 & 90.0 & 92.0 \\
 & \textit{Mean} & 97.1 & 3.6 & 14.6 & 25.5 & 95.2 & 95.1 \\
\midrule
32$\times$ & MK3 & 100.0 & 0.0 & 0.0 & 0.0 & 98.0 & 98.0 \\
 & MV & 100.0 & 0.0 & 2.0 & 0.0 & 100.0 & 100.0 \\
 & CWE & 98.4 & 0.0 & 40.8 & 46.0 & 90.4 & 87.8 \\
 & QA1 & 90.0 & 16.0 & 14.0 & 14.0 & 94.0 & 92.0 \\
 & \textit{Mean} & 97.1 & 4.0 & 14.2 & 15.0 & 95.6 & 94.5 \\
\bottomrule
\end{tabular}}
\par\vspace{1.2em}

\noindent\resizebox{\columnwidth}{!}{%
\begin{tabular}{llrrrrrr}
\toprule
\multicolumn{8}{c}{\textbf{Qwen3.5-9B, RULER-128K}} \\
\midrule
Rate & Task & Dense & KVz & F-KVz & PC+K & VKV+K & VKV+F \\
\midrule
8$\times$ & MK3 & 100.0 & 4.0 & 16.0 & 2.0 & 100.0 & 100.0 \\
 & MV & 100.0 & 46.5 & 4.5 & 2.0 & 100.0 & 100.0 \\
 & CWE & 100.0 & 47.0 & 59.2 & 84.4 & 97.6 & 95.4 \\
 & QA1 & 86.0 & 30.0 & 18.0 & 50.0 & 76.0 & 78.0 \\
 & \textit{Mean} & 96.5 & 31.9 & 24.4 & 34.6 & 93.4 & 93.3 \\
\midrule
16$\times$ & MK3 & 100.0 & 0.0 & 0.0 & 0.0 & 94.0 & 88.0 \\
 & MV & 100.0 & 3.0 & 2.0 & 0.0 & 100.0 & 100.0 \\
 & CWE & 100.0 & 6.2 & 48.0 & 74.8 & 87.6 & 93.0 \\
 & QA1 & 86.0 & 16.0 & 18.0 & 32.0 & 72.0 & 74.0 \\
 & \textit{Mean} & 96.5 & 6.3 & 17.0 & 26.7 & 88.4 & 88.8 \\
\midrule
32$\times$ & MK3 & 100.0 & 0.0 & 0.0 & 0.0 & 94.0 & 86.0 \\
 & MV & 100.0 & 0.0 & 2.0 & 0.0 & 100.0 & 100.0 \\
 & CWE & 100.0 & 0.0 & 47.4 & 45.0 & 79.0 & 73.6 \\
 & QA1 & 86.0 & 14.0 & 16.0 & 14.0 & 82.0 & 76.0 \\
 & \textit{Mean} & 96.5 & 3.5 & 16.4 & 14.8 & 88.8 & 83.9 \\
\bottomrule
\end{tabular}}
\par\vspace{1.2em}

\noindent\resizebox{\columnwidth}{!}{%
\begin{tabular}{llrrrrrr}
\toprule
\multicolumn{8}{c}{\textbf{Qwen3.5-9B, LOFT-32K}} \\
\midrule
Rate & Task & Dense & KVz & F-KVz & PC+K & VKV+K & VKV+F \\
\midrule
8$\times$ & NQ & 68.0 & 0.0 & 0.0 & 32.0 & 36.0 & 52.0 \\
 & HotpotQA & 60.0 & 16.0 & 8.0 & 44.0 & 60.0 & 64.0 \\
 & MuSiQue & 36.0 & 4.0 & 0.0 & 12.0 & 32.0 & 44.0 \\
 & QAMPARI & 64.0 & 0.0 & 0.0 & 0.0 & 56.0 & 60.0 \\
 & QUEST & 44.0 & 0.0 & 0.0 & 0.0 & 32.0 & 40.0 \\
 & \textit{Mean} & 54.4 & 4.0 & 1.6 & 17.6 & 43.2 & 52.0 \\
\midrule
16$\times$ & NQ & 68.0 & 0.0 & 12.0 & 0.0 & 24.0 & 24.0 \\
 & HotpotQA & 60.0 & 0.0 & 0.0 & 20.0 & 72.0 & 76.0 \\
 & MuSiQue & 36.0 & 0.0 & 0.0 & 0.0 & 36.0 & 24.0 \\
 & QAMPARI & 64.0 & 0.0 & 0.0 & 0.0 & 68.0 & 68.0 \\
 & QUEST & 44.0 & 0.0 & 0.0 & 0.0 & 40.0 & 40.0 \\
 & \textit{Mean} & 54.4 & 0.0 & 2.4 & 4.0 & 48.0 & 46.4 \\
\midrule
32$\times$ & NQ & 68.0 & 0.0 & 16.0 & 0.0 & 68.0 & 52.0 \\
 & HotpotQA & 60.0 & 0.0 & 0.0 & 0.0 & 68.0 & 56.0 \\
 & MuSiQue & 36.0 & 0.0 & 0.0 & 0.0 & 40.0 & 36.0 \\
 & QAMPARI & 64.0 & 0.0 & 0.0 & 0.0 & 68.0 & 60.0 \\
 & QUEST & 44.0 & 0.0 & 0.0 & 0.0 & 40.0 & 28.0 \\
 & \textit{Mean} & 54.4 & 0.0 & 3.2 & 0.0 & 56.8 & 46.4 \\
\bottomrule
\end{tabular}}
\par\vspace{1.2em}

\noindent\resizebox{\columnwidth}{!}{%
\begin{tabular}{llrrrrrr}
\toprule
\multicolumn{8}{c}{\textbf{Qwen3.5-9B, LOFT-128K}} \\
\midrule
Rate & Task & Dense & KVz & F-KVz & PC+K & VKV+K & VKV+F \\
\midrule
8$\times$ & NQ & 56.0 & 0.0 & 4.0 & 16.0 & 36.0 & 48.0 \\
 & HotpotQA & 64.0 & 0.0 & 24.0 & 44.0 & 52.0 & 56.0 \\
 & MuSiQue & 36.0 & 0.0 & 4.0 & 0.0 & 24.0 & 20.0 \\
 & QAMPARI & 28.0 & 0.0 & 0.0 & 0.0 & 40.0 & 16.0 \\
 & QUEST & 36.0 & 0.0 & 4.0 & 4.0 & 20.0 & 16.0 \\
 & \textit{Mean} & 44.0 & 0.0 & 7.2 & 12.8 & 34.4 & 31.2 \\
\midrule
16$\times$ & NQ & 56.0 & 0.0 & 4.0 & 4.0 & 44.0 & 52.0 \\
 & HotpotQA & 64.0 & 0.0 & 4.0 & 20.0 & 52.0 & 44.0 \\
 & MuSiQue & 36.0 & 0.0 & 4.0 & 0.0 & 24.0 & 32.0 \\
 & QAMPARI & 28.0 & 0.0 & 0.0 & 0.0 & 32.0 & 4.0 \\
 & QUEST & 36.0 & 0.0 & 0.0 & 0.0 & 28.0 & 12.0 \\
 & \textit{Mean} & 44.0 & 0.0 & 2.4 & 4.8 & 36.0 & 28.8 \\
\midrule
32$\times$ & NQ & 56.0 & 0.0 & 4.0 & 0.0 & 48.0 & 32.0 \\
 & HotpotQA & 64.0 & 0.0 & 4.0 & 0.0 & 64.0 & 56.0 \\
 & MuSiQue & 36.0 & 0.0 & 4.0 & 0.0 & 28.0 & 32.0 \\
 & QAMPARI & 28.0 & 0.0 & 0.0 & 0.0 & 4.0 & 20.0 \\
 & QUEST & 36.0 & 0.0 & 0.0 & 0.0 & 12.0 & 12.0 \\
 & \textit{Mean} & 44.0 & 0.0 & 2.4 & 0.0 & 31.2 & 30.4 \\
\bottomrule
\end{tabular}}
\par\vspace{1.2em}

\noindent\resizebox{\columnwidth}{!}{%
\begin{tabular}{llrrrrrr}
\toprule
\multicolumn{8}{c}{\textbf{Ministral-3-8B, RULER-32K}} \\
\midrule
Rate & Task & Dense & KVz & F-KVz & PC+K & VKV+K & VKV+F \\
\midrule
8$\times$ & MK3 & 100.0 & 28.0 & 68.0 & 100.0 & 98.0 & 100.0 \\
 & MV & 98.0 & 92.0 & 86.0 & 98.0 & 95.5 & 94.0 \\
 & CWE & 76.4 & 40.0 & 31.8 & 74.6 & 75.2 & 79.4 \\
 & QA1 & 74.0 & 60.0 & 62.0 & 62.0 & 68.0 & 66.0 \\
 & \textit{Mean} & 87.1 & 55.0 & 62.0 & 83.7 & 84.2 & 84.8 \\
\midrule
16$\times$ & MK3 & 100.0 & 14.0 & 16.0 & 86.0 & 98.0 & 100.0 \\
 & MV & 98.0 & 54.5 & 38.0 & 57.0 & 99.5 & 95.5 \\
 & CWE & 76.4 & 13.0 & 26.8 & 65.6 & 80.4 & 76.0 \\
 & QA1 & 74.0 & 42.0 & 44.0 & 60.0 & 72.0 & 74.0 \\
 & \textit{Mean} & 87.1 & 30.9 & 31.2 & 67.2 & 87.5 & 86.4 \\
\midrule
32$\times$ & MK3 & 100.0 & 0.0 & 0.0 & 0.0 & 88.0 & 94.0 \\
 & MV & 98.0 & 26.0 & 3.0 & 2.0 & 96.0 & 98.0 \\
 & CWE & 76.4 & 1.8 & 12.6 & 47.2 & 76.2 & 74.8 \\
 & QA1 & 74.0 & 20.0 & 16.0 & 40.0 & 74.0 & 74.0 \\
 & \textit{Mean} & 87.1 & 11.9 & 7.9 & 22.3 & 83.5 & 85.2 \\
\bottomrule
\end{tabular}}
\par\vspace{1.2em}

\noindent\resizebox{\columnwidth}{!}{%
\begin{tabular}{llrrrrrr}
\toprule
\multicolumn{8}{c}{\textbf{Ministral-3-8B, RULER-128K}} \\
\midrule
Rate & Task & Dense & KVz & F-KVz & PC+K & VKV+K & VKV+F \\
\midrule
8$\times$ & MK3 & 100.0 & 0.0 & 20.0 & 95.0 & 94.0 & 88.0 \\
 & MV & 97.5 & 81.2 & 77.5 & 96.2 & 99.0 & 96.0 \\
 & CWE & 10.0 & 6.5 & 11.0 & 7.0 & 8.0 & 7.2 \\
 & QA1 & 80.0 & 40.0 & 75.0 & 80.0 & 60.0 & 68.0 \\
 & \textit{Mean} & 71.9 & 31.9 & 45.9 & 69.6 & 65.2 & 64.8 \\
\midrule
16$\times$ & MK3 & 100.0 & 0.0 & 10.0 & 55.0 & 70.0 & 64.0 \\
 & MV & 97.5 & 50.0 & 37.5 & 51.2 & 97.0 & 96.0 \\
 & CWE & 10.0 & 0.0 & 3.5 & 6.5 & 7.2 & 7.2 \\
 & QA1 & 80.0 & 30.0 & 45.0 & 65.0 & 58.0 & 56.0 \\
 & \textit{Mean} & 71.9 & 20.0 & 24.0 & 44.4 & 58.0 & 55.8 \\
\midrule
32$\times$ & MK3 & 100.0 & 0.0 & 0.0 & 0.0 & 30.0 & 30.0 \\
 & MV & 97.5 & 5.0 & 0.0 & 5.0 & 94.5 & 91.0 \\
 & CWE & 10.0 & 0.5 & 4.5 & 4.0 & 7.0 & 8.2 \\
 & QA1 & 80.0 & 20.0 & 30.0 & 40.0 & 56.0 & 60.0 \\
 & \textit{Mean} & 71.9 & 6.4 & 8.6 & 12.2 & 46.9 & 47.3 \\
\bottomrule
\end{tabular}}
\par\vspace{1.2em}

\noindent\resizebox{\columnwidth}{!}{%
\begin{tabular}{llrrrrrr}
\toprule
\multicolumn{8}{c}{\textbf{Ministral-3-8B, LOFT-32K}} \\
\midrule
Rate & Task & Dense & KVz & F-KVz & PC+K & VKV+K & VKV+F \\
\midrule
8$\times$ & NQ & 76.0 & 32.0 & 52.0 & 72.0 & 76.0 & 72.0 \\
 & HotpotQA & 68.0 & 56.0 & 56.0 & 48.0 & 72.0 & 72.0 \\
 & MuSiQue & 32.0 & 40.0 & 36.0 & 40.0 & 36.0 & 28.0 \\
 & QAMPARI & 36.0 & 8.0 & 0.0 & 28.0 & 40.0 & 48.0 \\
 & QUEST & 36.0 & 12.0 & 8.0 & 32.0 & 40.0 & 36.0 \\
 & \textit{Mean} & 49.6 & 29.6 & 30.4 & 44.0 & 52.8 & 51.2 \\
\midrule
16$\times$ & NQ & 76.0 & 8.0 & 28.0 & 44.0 & 64.0 & 64.0 \\
 & HotpotQA & 68.0 & 52.0 & 40.0 & 48.0 & 64.0 & 52.0 \\
 & MuSiQue & 32.0 & 4.0 & 12.0 & 16.0 & 36.0 & 40.0 \\
 & QAMPARI & 36.0 & 0.0 & 0.0 & 4.0 & 24.0 & 48.0 \\
 & QUEST & 36.0 & 16.0 & 0.0 & 24.0 & 20.0 & 28.0 \\
 & \textit{Mean} & 49.6 & 16.0 & 16.0 & 27.2 & 41.6 & 46.4 \\
\midrule
32$\times$ & NQ & 76.0 & 0.0 & 4.0 & 24.0 & 72.0 & 56.0 \\
 & HotpotQA & 68.0 & 0.0 & 8.0 & 16.0 & 64.0 & 68.0 \\
 & MuSiQue & 32.0 & 0.0 & 0.0 & 12.0 & 32.0 & 28.0 \\
 & QAMPARI & 36.0 & 0.0 & 0.0 & 0.0 & 36.0 & 32.0 \\
 & QUEST & 36.0 & 0.0 & 0.0 & 12.0 & 24.0 & 20.0 \\
 & \textit{Mean} & 49.6 & 0.0 & 2.4 & 12.8 & 45.6 & 40.8 \\
\bottomrule
\end{tabular}}
\par\vspace{1.2em}

\noindent\resizebox{\columnwidth}{!}{%
\begin{tabular}{llrrrrrr}
\toprule
\multicolumn{8}{c}{\textbf{Ministral-3-8B, LOFT-128K}} \\
\midrule
Rate & Task & Dense & KVz & F-KVz & PC+K & VKV+K & VKV+F \\
\midrule
8$\times$ & NQ & 48.0 & 20.0 & 20.0 & 68.0 & 36.0 & 52.0 \\
 & HotpotQA & 64.0 & 36.0 & 36.0 & 64.0 & 68.0 & 64.0 \\
 & MuSiQue & 52.0 & 40.0 & 52.0 & 12.0 & 40.0 & 48.0 \\
 & QAMPARI & 12.0 & 0.0 & 4.0 & 16.0 & 24.0 & 12.0 \\
 & QUEST & 28.0 & 4.0 & 12.0 & 24.0 & 24.0 & 32.0 \\
 & \textit{Mean} & 40.8 & 20.0 & 24.8 & 36.8 & 38.4 & 41.6 \\
\midrule
16$\times$ & NQ & 48.0 & 4.0 & 28.0 & 36.0 & 68.0 & 36.0 \\
 & HotpotQA & 64.0 & 24.0 & 24.0 & 40.0 & 60.0 & 48.0 \\
 & MuSiQue & 52.0 & 12.0 & 36.0 & 4.0 & 24.0 & 32.0 \\
 & QAMPARI & 12.0 & 0.0 & 0.0 & 4.0 & 12.0 & 12.0 \\
 & QUEST & 28.0 & 0.0 & 0.0 & 8.0 & 16.0 & 16.0 \\
 & \textit{Mean} & 40.8 & 8.0 & 17.6 & 18.4 & 36.0 & 28.8 \\
\midrule
32$\times$ & NQ & 48.0 & 0.0 & 8.0 & 16.0 & 40.0 & 48.0 \\
 & HotpotQA & 64.0 & 0.0 & 24.0 & 0.0 & 44.0 & 44.0 \\
 & MuSiQue & 52.0 & 0.0 & 12.0 & 12.0 & 24.0 & 28.0 \\
 & QAMPARI & 12.0 & 0.0 & 0.0 & 0.0 & 12.0 & 8.0 \\
 & QUEST & 28.0 & 0.0 & 0.0 & 0.0 & 8.0 & 20.0 \\
 & \textit{Mean} & 40.8 & 0.0 & 8.8 & 5.6 & 25.6 & 29.6 \\
\bottomrule
\end{tabular}}
\par\vspace{1.2em}